\documentclass{article}

\usepackage[left=2.5cm, right=2.5cm]{geometry}

\usepackage{times}
\usepackage{soul}
\usepackage{url}
\usepackage[hidelinks]{hyperref}
\usepackage[utf8]{inputenc}
\usepackage[small]{caption}
\usepackage{graphicx}
\usepackage{amsmath}
\usepackage{booktabs}
\usepackage{multirow}
\usepackage{multicol}
\usepackage{tabularx}
\usepackage{csquotes}

\usepackage[square,numbers]{natbib}
\usepackage{graphicx}

\usepackage{amsmath,amsfonts,bm,amssymb,amsthm}
\usepackage{mathtools}
\usepackage[capitalize]{cleveref}
\usepackage{subcaption}
\usepackage[dvipsnames]{xcolor}
\usepackage{tikz}

\usepackage{thmtools,thm-restate}
\usepackage{todonotes}

\usetikzlibrary{patterns,snakes}
\def\layersep{1.8cm}
\usetikzlibrary{arrows.meta}

\hypersetup{
    citecolor=blue,
    colorlinks=true,
    linkcolor=blue,
    filecolor=blue,
    urlcolor=blue,
}

\usepackage{definitions}

\title{
  Empirical Evaluation and Theoretical Analysis \\
  for Representation Learning: A Survey
}

\author{
Kento Nozawa$^{1,2}$
Issei Sato$^1$\\
$^1$The University of Tokyo, $^2$RIKEN AIP\\
\{nzw, sato\}@g.ecc.u-tokyo.ac.jp
}

\begin{document}

\maketitle

\begin{abstract}
  Representation learning enables us to automatically extract generic feature representations from a dataset to solve another machine learning task.
  Recently, extracted feature representations by a representation learning algorithm and a simple predictor have exhibited state-of-the-art performance on several machine learning tasks.
  Despite its remarkable progress, there exist various ways to evaluate representation learning algorithms depending on the application because of the flexibility of representation learning.
  To understand the current representation learning, we review evaluation methods of representation learning algorithms and theoretical analyses.
  On the basis of our evaluation survey, we also discuss the future direction of representation learning.
  Note that this survey is the extended version of \citet{Nozawa2022IJCAI-ECAI}.
\end{abstract}

\setcounter{tocdepth}{2}
\tableofcontents

\section{Introduction}
\label{sec:introduction}

Deep learning~\citep{Goodfellow-et-al-2016} is a major workhorse in modern machine learning algorithms for image recognition~\citep{Krizhevsky2012NeurIPS}, semantic segmentation~\citep{Ronneberger2015MICCAI}, generative models~\citep{Kingma2014ICLR,Goodfellow2014NeurIPS}, machine translation~\citep{Vaswani2017NeurIPS}, speech recognition~\citep{Hinton2012IEEESPM}, reinforcement learning~\citep{Mnih2015Nature}, and so on.
An attractive nature of deep learning is expressiveness, such that deep neural networks approximate a complex function to solve machine learning tasks.
Thanks to the expressiveness of deep neural networks, a recent high-performance computing device, e.g., general-purpose computing with graphics processing units (GPGPUs) or tensor processing unit (TPU),%
\footnote{\url{https://cloud.google.com/tpu}}
and a massive training dataset,
deep learning-based algorithms can outperform human performance on multiple machine learning benchmark tasks, for example,
ResNet-101~\citep{He2016CVPR} on the ImageNet object categorization dataset~\citep{Deng2009CVPR} and DeBERTa~\citep{He2021ICLR} on SuperGLUE~\citep{Wang2019NeurIPSa} that is a set of benchmark tasks for natural language processing.

In deep neural networks, multiple nonlinear transformations from input space to output space are distinguished characteristics compared with other machine learning algorithms such as a kernel method~\citep{Shawe-Taylor2004book}.
Nonlinear transformations enable deep neural networks to internally learn a feature vector, namely, feature ``representation'', that effectively captures informative features to optimize the objective function.
For example, when we solve a digit classification task with deep learning, namely, MNIST,
the input image is transformed to a more abstract representation than the original input to predict its class label, which is a digit, after applying multiple nonlinear transformations by using convolutional neural networks~\citep{LeCun1998IEEE}.
Thanks to this nonlinearity, deep learning algorithms often lower the priority of feature engineering.
In other words, we require much less domain knowledge to carefully construct hand-crafted features when we solve the machine learning problem.
For example, the winning team of ILSVRC2012%
\footnote{The competition results are available at \url{http://www.image-net.org/challenges/LSVRC/2012/}.}
used a deep convolutional network called AlexNet~\citep{Krizhevsky2012NeurIPS} that can take raw pixels of an image to predict a class label.
On the other hand, the 2nd, 3rd, and 4th place teams created feature vectors including Improved Fisher vector~\citep{Perronnin2010ECCV} with local descriptors such as scale-invariant feature transform (SIFT)~\citep{Lowe2004IJCV} as an input feature vector of a support vector machine.
In this competition, learning internal representations using deep learning played a critical role in being superior to hand-crafted feature representations.
Deep neural network-based algorithms have been dominant in the task since this competition.

Motivated by the importance of learning feature representations,
representation learning%
\footnote{We use representation learning and feature learning, interchangeably.}
is defined as a set of methods that automatically learn discriminative feature representations from a dataset to solve a machine learning task~\citep{LeCun2015Nature}.
Empirically, the learned model is used as a feature extractor for other machine learning tasks, such as classification, regression, and visualization.
In this sense, representation learning is also referred to as method to learn generic feature representations for \textit{unseen} downstream tasks rather than end-to-end methods to solve a machine learning task directly.
Unfortunately, we do not yet have a well-defined evaluation metric of representation learning for the latter case because of various applications of representation learning.
Nevertheless, we believe that evaluation methods are critical in designing novel or analyzing existing algorithms.

We review the existing evaluation methods of representation learning algorithms to understand their applications and the common practice.
Specifically, we propose four evaluation perspectives of representation learning algorithms.
In addition, we review theoretical analyses on representation learning algorithms in \cref{sec:theoretical-analysis}.
Note that we \textit{do not} aim to provide a comprehensive survey on the state-of-the-art algorithms compared with existing representation learning surveys discussed in \cref{sec:related-work}.

\section{Related Work}
\label{sec:related-work}
In this section, we discuss the existing surveys and related papers.

\subsection{Surveys on General Representation Learning}
\citet{Bengio2013IEEE} provided the first review on unsupervised representation learning and deep learning.
They discussed the importance of representation learning and learning algorithms such as auto-encoder and deep brief networks (DBNs).
\citet{Zhong2016JFDS} reviewed the historical developments of feature learning, especially non-deep learning methods and deep learning methods.
\citet{Zhong2016JFDS} mainly explained the formulations of non-deep learning-based feature learning algorithms.
Compared with these survey papers, we will review representation learning from evaluation perspectives.

\subsection{Surveys on Specific Representation Learning}
Depending on the target domains that we applied representation learning to, there exist comprehensive surveys on representation learning to review the state-of-the-art algorithms.
Researchers also survey specific types of representation learning algorithms, such as self-supervised representation learning, as mentioned later.
Compared with them, this paper does not review state-of-the-art algorithms due to their too rapid progress.
Instead, we focus on a more general perspective: the evaluation of representation learning algorithms and theoretical analyses.

\paragraph{Graph}
Graph representation learning aims to learn feature representations for a node, edge, subgraph, or whole graph depending on a downstream task.
Graph representation learning algorithms have been actively proposed since the emergence of ``word2vec''~\citep{Mikolov2013ICLR,Mikolov2013NeurIPS} for learning word representations.
Pioneers of node representation learning~\citep{Perozzi2014KDD,Tang2015WWW} borrow techniques from word2vec.
We refer to comprehensive review papers~\citep{Hamilton2017IEEE,Goyal2018KBS,Cai2018IEEETKD,Chen2020APSIPA,Zhang2020IEEE,Kazemi2020JMLR} and the book~\citep{Hamilton2020book} for more technical details about graph representation learning.
\citet{Kirong2021arXiv} reviewed a specific type of graph representation learning algorithm: contrastive representation learning.
Even though our review mainly focuses on vision or language domain as an example,
our explanations and evaluation procedures are applicable to graph representation learning as well.

\paragraph{Knowledge Graph}
Knowledge graph representation learning is an active subfield of graph representation learning.
Knowledge graph representation learning embeds the knowledge graph's entity, such as ``London'' or ``England''; or the relation of entities, such as ``capital of'', into lower dimensional space to expand the knowledge graph by predicting a missing entity or relation.
We refer to comprehensive review papers~\citep{Wang2017IEEETKDE,Lin2018arXiv,Ji2021IEEE} for more details about knowledge graph representation learning.

\paragraph{Natural Language Processing}
In natural language processing (NLP), neural network-based algorithms often use vector representations for a minimal unit, such as a word or character;
learning representations play an important role in the algorithms or downstream tasks.
\citet{Smith2020CACM} gave a seminal review on the historical progress of word representations in NLP.
Recent textbooks of NLP cover representation learning as a chapter~\citep[Chapter 14]{Eisenstein2019Book} or the whole~\citep{Liu2020bBook}.
\citet{Rethmeier2021arXiv} reviewed a recent representation learning approach: contrastive pre-training.
\citet{Rogers2020TACL,Xia2020EMNLP} surveyed a specific representation learning algorithm: BERT~\citep{Devlin2019NAACL} and its variants.

There exist algorithm-specific papers for evaluation methods of unsupervised word embeddings~\citep{Schnabel2015EMNLP} and probabilistic topic models~\citep{Wallach2009ICML}.
These papers explained detailed evaluation methods of the target algorithms and showed empirical results.
In contrast, we will consider algorithm-agnostic evaluation procedures to give a broad view of representation learning.

\paragraph{Multi-modality}
When we access multiple modalities of data such as news articles (image and text) or YouTube videos (sequence of images, audio, and text description),
we can design a unified algorithm by combining representation learning algorithms for each unimodality.
Such representation learning is called multi-modal representation learning.
\citet{Guo2019Access} provided a comprehensive survey on deep multi-modal representation learning.
\citet{Li2019KDE} provided an overview of multiview representation learning, which is a more general setting than a multi-modal one.
Note that this direction has been attracting attention as a form of weak supervision for the last few years thanks to unified neural network architecture, as demonstrated by CLIP~\citep{Radford2021ICML}.

\paragraph{Self-supervised Representation Learning}
Last few years, self-supervised representation learning algorithms, especially contrastive representation learning, have been attracting much attention from the machine learning community because the learned representation yield informative feature representations for downstream tasks in practice.
Self-supervised representation learning attempts to learn generic feature representations from only an unlabeled dataset.
\citet{Le-Khac2020IEEEAccess,Jaiswal2021T} gave an overview of state-of-the-art contrastive representation learning algorithms.
\citet{Jing2021TPAMI} reviewed self-supervised learning algorithms and pretext tasks in the vision domain and compared algorithms as a feature extractor for downstream classification tasks.
\citet{Schmarje2021IEEEAccess} organized common concepts used in self-supervised learning and related frameworks such as supervised, semi-supervised, and unsupervised learning for image classification.
\citet{Schmarje2021IEEEAccess} also compared representation learning algorithms on the basis of classification performance and common ideas.
For a specific downstream task, \citet{Huang2021ArXiv211014711Cs} reviewed a self-supervised learning approach for few-shot object detection.
As another resource to know the advance of self-supervised learning research, we refer the readers to a tutorial at NeurIPS 2021 by~\citet{Weng2021NeurIPStutorial}.
Mainly these survey papers focused on the state-of-the-art representation learning algorithms and provided a taxonomy to understand self-supervised representation learning.

\subsection{Related Machine Learning Formulation}
Representation learning shares common concepts with other machine learning formulations.
Here, we briefly mention the difference between representation learning and them.
We also refer to these survey papers as pointers.

\paragraph{Semi-supervised learning}
Semi-supervised learning~\citep{Chapelle2006Book} trains a predictor as accurately as possible on many unlabeled data and few labeled data.
We can evaluate the effectiveness of a representation learning algorithm as a semi-supervised learner.
\cref{sec:pre-training,sec:auxiliary} discuss how to apply representation learning algorithms to the semi-supervised learning problem.

\paragraph{Transfer Learning}
Transfer learning~\citep{Pan2010IEEE,Weiss2016BigData} aims to improve a target task's performance using knowledge obtained from a source task that differs from the target task.
We refer to~\citet{Redko2020} as a survey of ``domain adaptation'', which is one of transfer learning formulations and its theory.
Similar to semi-supervised learning, we can evaluate a representation learning algorithm as a transfer learning algorithm.
For example, we train a feature extractor on a source task and train the learned feature extractor with an additional predictor on a target task as explained in~\cref{sec:pre-training}.
\cref{sec:regularization,sec:auxiliary} will also mention how to apply representation learning to transfer learning.

\paragraph{Metric Learning}
Metric learning~\citep{Kulis2012,Bellet2014} concerns learning a model that calculates the similarity between data samples.
Metric learning maps input samples into a metric space such that samples with the same labels are similar in terms of a pre-defined metric such as cosine similarity, and the samples with different labels are dissimilar.
Metric learning and representation learning map an input sample into a feature space.
In addition, they use a similar objective function, for example,
$N$-pair loss~\citep{Sohn2016NeurIPS} for metric learning and InfoNCE loss~\citep{Oord2018arXiv} for contrastive representation learning.
The difference between metric learning and representation learning is that metric learning focuses on similarity between data samples.
On the other hand, representation learning focuses on learning generic feature representations.

\section{Background: Representation Learning}
\label{sec:representation-learning}

We give a high-level overview and formulation of representation learning.
We explain two formulations in terms of the existence of supervised signals during representation learning:
supervised representation learning (\cref{sec:sup-repre-learning}) and unsupervised representation learning (\cref{sec:unsup-repre-learning}).

\subsection{Supervised Representation Learning}
\label{sec:sup-repre-learning}

Suppose supervised dataset $\Dcal_\mathrm{sup} = \{ (\xbf_i, y_i) \}_{i=1}^N$, where $\xbf$ is an input sample, and $y$ is a supervised signal such as a class label in classification or a real-valued target vector in regression.
For example, $\xbf \in \Rbb^{\mathrm{channel} \times \mathrm{height} \times \mathrm{width}}$ for the color image domain;
$\xbf \in \Rbb^{T \times  \mathrm{channel} \times \mathrm{height} \times \mathrm{width}}$, where $T$ is the length of time-steps, for the video domain;
and $\xbf = [ \obf_1, \ldots, \obf_T] $ represents a sequence of words and each $\obf_t$ is a one-hot vector of the  $t^\mathrm{th}$ word for natural language processing.
As a running example, we suppose classification, where $\xbf$ is an input sample, and $y \in \Ycal = [1, \ldots, Y]$ is a categorical value in pre-defined class set $\Ycal$.

A supervised representation learning algorithm trains parameterized feature extractor $\hbf$ by solving a supervised task on $\Dcal_\mathrm{sup}$.
Feature extractor $\hbf: \Rbb^I \rightarrow \Rbb^d$ maps an input representation $\xbf$ to a feature representation $\hbf(\xbf) \in \Rbb^d$, where $d$ tends to be smaller than $I$, the dimensionality of $\xbf$.
Depending on the formulation of the supervised task, an additional function, $\gbf: \Rbb^d \rightarrow \Rbb^O$, yields the output representation to evaluate a supervised objective function given feature representation $\hbf(\xbf)$.
For multi-class classification, output representation is a real-valued vector, where $O = Y$, when we use a softmax cross-entropy loss.
Formally, we minimize training loss function $\widehat{L}(\hbf, \gbf)$ instantiated as softmax cross-entropy loss to obtain pre-trained $\widehat{\hbf}$ and $\widehat{\gbf}$ as follows:
\begin{align}
  \widehat{\hbf}, \hspace{1ex} \widehat{\gbf} &= \argmin_{\hbf, \gbf} \widehat{L}(\hbf, \gbf),
  \nonumber \\
  \text{ where }
  \hspace{1ex}
  \widehat{L}(\hbf, \gbf) & = - \frac{1}{N} \sum_{i=1}^{N} \ln \frac{ \exp(\gbf_{y_i}(\hbf(\xbf_i))) }{ {\displaystyle \sum_{y \in \Ycal} \exp( \gbf_{y}(\hbf(\xbf_i))) } }.
  \label{eq:supervised-loss}
\end{align}
After minimizing supervised loss $\widehat{L}$ \eqref{eq:supervised-loss}, we use $\widehat{\hbf}$ as a feature extractor for other machine learning tasks.
We tend to remove the rest part, $\widehat{\gbf}$ because representation $\widehat{\gbf}(\widehat{\hbf}(\xbf))$ leads to poor downstream performance in practice~\citep{Donahue2014ICML}.
\citet{Yosinski2014NeurIPS} explained that feature representations extracted by using near the final layer in neural networks are too specialized to solve the upstream supervised task without fine-tuning.
Similar results have been reported in unsupervised representation learning~\citep{Hjelm2019ICLR,Bachman2019NeurIPS,Baevski2021NeurIPS}.
Following the notations above, we define supervised representation learning as follows:

\begin{definition}[Supervised Representation Learning]
  \label{definition:supervised-representation-learning}
  Supervised representation learning aims to learn generic feature extractor $\hbf: \Rbb^{I} \rightarrow \Rbb^d$ by optimizing $\widehat{L}$ on labeled dataset $\Dcal_\mathrm{sup}$ automatically without feature engineering by domain experts.
\end{definition}

Supervised learning algorithms can be viewed as supervised representation learning.
For example, neural networks trained on ImageNet~\citep{Deng2009CVPR} or Youtube-8M~\citep{Abu-El-Haija2016arXiv}, and machine translation models trained on a parallel corpus of English and French.%
\footnote{For example, WMT2014 dataset at \url{http://www.statmt.org/wmt14/}.}
Concretely, DeCAF~\citep{Donahue2014ICML} formulated ImageNet classification as a representation learning task and demonstrated the effectiveness of the learned feature extractor for downstream tasks.
In NLP, \citet{Conneau2017EMNLP} proposed solving the Stanford natural language inference (NLI) task~\citep{Bowman2015EMNLP} as a representation learning task, and \citet{McCann2017NeurIPS} proposed solving a machine translation task to train a sentence feature extractor.

One of the advantages of supervised representation learning is that we obtain feature extractor $\widehat{\hbf}$ as a by-product of supervised learning.
For example, VGG~\citep{Simonyan2015ICLR} was originally proposed for the ImageNet classification task, nevertheless, the trained models have been widely used as a pre-trained feature extractor for other vision tasks such as segmentation~\citep{Long2015CVPR};
and object detection~\citep{Girshick2014CVPR,Girshick2015ICCV,Ren2015NeurIPS}.
Indeed, we can download many pre-trained models proposed for well-defined tasks from the authors' project pages or hub sites.%
\footnote{For example, \url{https://huggingface.co/models}, \url{https://tfhub.dev}, and \url{https://pytorch.org/hub}.}

Empirically large sample size in supervised representation learning improves downstream performance~\citep{Conneau2017EMNLP,Sun2017ICCV,Mahajan2018ECCV,Kolesnikov2020ECCV}.
Unfortunately, enlarging the sample size is costly regarding time and money by hiring annotators and teaching them how to annotate data.
In addition, we expect that the supervised task is not too easy to capture generic representations for downstream tasks.
Intuitively, if we pre-train a model on a difficult task such as ImageNet classification, the model can generalize well to a simpler task, such as MNIST classification.
However, the reverse probably does not hold; the pre-trained model on MNIST does not generalize well to ImageNet because the model trained on MNIST could not see complicated patterns during the training to solve ImageNet classification.%
\footnote{As demonstrated by~\citet{Kataoka2020ACCV}, fine-tuning a supervised pre-train model on synthetic images might reduce the necessity of such a difficult task.}
Creating a dataset for a difficult task, which is ImageNet in the example above, does not only require skilled annotators but also easily contaminates the dataset that could hurt upstream performance~\citep{Beyer2020arXiv}.
As a result, the pre-trained model performs poorly as a feature extractor for downstream tasks.
To overcome this disadvantage, unsupervised representation learning or weakly supervised representation learning~\citep{Radford2021ICML,Mahajan2018ECCV} have been attracting much attention from the machine learning community.
Shortly they replace the supervised label with auxiliary information such as user-provided tags~\citep{Mahajan2018ECCV} or a description~\citep{Radford2021ICML} in the supervised loss of an image classification task.
We do not introduce the weakly supervised representation learning in this review for simplicity.

\citet{Kornblith2019CVPR,Abnar2022ICLR} have reported that we can often predict the performance of downstream tasks by using the generalization performance in ImageNet when using ImageNet pre-training.
However, for even supervised representation learning, the best-performed model does not give the best performance on multiple downstream tasks~\citep{Abnar2022ICLR}.
A similar tendency has been reported in unsupervised representation learning such as~\citet{Ericsson2021CVPR}.

\subsection{Unsupervised Representation Learning}
\label{sec:unsup-repre-learning}

Unsupervised representation learning%
\footnote{We use unsupervised representation learning and self-supervised representation learning interchangeably.}
does not use label information at all to learn feature extractor $\hbf$.
Suppose unlabeled dataset $\Dcal_\mathrm{un} = \{ \xbf_i \}_{i=1}^M$, where $M$ is the number of unlabeled samples.
The difference from supervised representation learning is that unsupervised representation learning trains feature extractor $\hbf$ by solving an \textit {unsupervised} task on $\Dcal_\mathrm{un}$.
To do so, unsupervised representation learning papers~\citep{Jing2021TPAMI} proposed a novel unsupervised loss function that is called ``pretext task'' to replace a supervised loss function.
For example, auto-encoders~\citep{Rumelhart1986Chapter} minimize a reconstruction error as unsupervised loss $\widehat{L}_\mathrm{un}$ defined by
\begin{align}
  \widehat{L}_{\mathrm{un}}(\hbf, \gbf) &= \frac{1}{M} \sum_{i=1}^{M}
  \left\| \gbf(\hbf(\xbf_i)) - \xbf_i \right\|_2.
  \label{eq:auto-encoder-loss}
\end{align}
Intuitively, $\hbf$ compresses $\xbf$ such that $\gbf$ recovers $\xbf$ from $\hbf(\xbf)$ by minimizing \cref{eq:auto-encoder-loss}.
We expect that such compressed feature representation $\hbf(\xbf)$ captures useful features of $\xbf$ to solve other machine learning tasks.
As the counterpart of \cref{definition:supervised-representation-learning}, we define unsupervised representation learning as follows:
\begin{definition}[Unsupervised Representation Learning]
  \label{definition:unsupervised-representation-learning}
  Unsupervised representation learning aims to learn generic feature extractor $\hbf:  \Rbb^I \rightarrow \Rbb^d$ by optimizing $\widehat{L}_{\mathrm{un}}$ on unlabeled dataset $\Dcal_\mathrm{un}$ automatically without feature engineering by domain experts.
\end{definition}

Thanks to the unsupervised nature, we can easily increase the sample size of $D_\mathrm{un}$ at almost no cost.
For example, \citet{Mikolov2018LREC} trained word representations on $630$ billion words collected from \texttt{CommonCrawl}%
\footnote{Web crawl data at \url{https://commoncrawl.org}},
\citet{Du2021arXiv} trained from $1.6$ trillion tokens collected from multiple text datasets, and \citet{He2020CVPR,Caron2020NeurIPS,Goyal2021arXiv} trained self-supervised models on one billion images collected from \texttt{Instagram}.%
\footnote{Social networking service for sharing photos and videos, \url{https://www.instagram.com}.}
This property is desirable because enlarging the size of the dataset improves the performance of downstream tasks in practice~\citep{Zhai2020arXiv,Hernandez2021arXiv}.
Surprisingly, even if we train a feature extractor on the same amount of data, unsupervised representation learning gives better transfer performance~\citep{Ericsson2021CVPR,Kotar2021ICCV} and better generalization on out-of-distribution~\citep{Sariyildiz2021ICCV,Mitrovic2021ICLR} than supervised representation learning depending on downstream tasks.

One of the disadvantages of unsupervised representation learning is the difficulty of evaluation at the representation phase.
We do not even know the existence of a universal unsupervised objective that indicates the minimizer can guarantee downstream performance.
As an empirical observation, \citet{Kolesnikov2019CVPR} reported that lower validation loss of representation learning tasks did not imply better validation accuracy on ImageNet classification across different models.
Hence, the generalization performance of a downstream task is often used as an evaluation metric in practice, as explained in~\cref{sec:pre-training}.

\subsection{Type of Representations}
\label{sec:type-representation}

To emphasize the advantage of representation learning, we discuss the difference between local and distributed representations that are related to hand-crafted feature representations and learned feature representations, respectively.

Local representation~\citep{Hinton1986BookChapter} is one of the most common forms of feature representation created by domain experts.
We refer to a vector as a feature representation for a running example.
Local representation assigns one concrete attribute to one element in a vector, e.g., one-hot encoding for discrete data.
The one-hot vector's dimensionality is the number of pre-defined units.
Each data sample corresponds to each element one-to-one; if the one-hot vector's $i^{\mathrm{th}}$ element is one, it represents the $i^{\mathrm{th}}$ sample.
For example, when we treat a word ``game'' as a one-hot representation with a pre-defined vocabulary whose size is $V$, the local representation of cat $\xbf_{\text{game}}$ is defined by
\begin{align}
  \xbf_{\text{game}} =
  [\overbrace
  {\underbrace{0}_{\text{the}}, \underbrace{0}_{\text{imitation}}, \underbrace{1}_{\text{game}}, \ldots, \underbrace{0}_{\text{think}}}
  ^{V}].
  \label{eq:one-hot-vec}
\end{align}
Likewise, a frequency vector can represent a multi-set of discrete data.
For example, each element in a vector corresponds to the word frequencies to represent the phrase ``the imitation game''\footnote{The title of Section 1 of \citet{Turing1950Mind}.}:
\begin{align}
  \xbf_{\text{the imitation game}} =
  [\overbrace
    {\underbrace{1}_{\text{the}}, \underbrace{1}_{\text{imitation}}, \underbrace{1}_{\text{game}}, \ldots, \underbrace{0}_{\text{think}}}
    ^{V}
  ].
  \label{eq:freq-vec}
\end{align}
Another example is a hand-crafted feature vector.
For example, the input sample of the Iris dataset~\citep{Fisher1936AE} is in $\Rbb^4$, where each element in a vector is defined as follows:
\begin{align}
  \xbf =
    [\underbrace{5.1}_{\text{sepal length}}, \underbrace{3.5}_{\text{sepal width}}, \underbrace{1.4}_{\text{petal length}}, \underbrace{0.2}_{\text{petal width}}].
    \label{eq:iris-vec}
\end{align}
These local representations are interpretable because each element in a vector has a clear meaning.
However, they have a disadvantage in capturing complicated features.
For example, for frequency vector representation \eqref{eq:freq-vec}, the representation loses the order of words because this representation only uses word frequencies.
To deal with the order of words in the sentence, we might add $n$-gram counts;
however, it causes the curse of dimensionality~\citep{Bengio2003JMLR}.
In addition, creating local representations by domain experts is costly to apply machine learning to various real-world problems.

Learning representation automatically from a dataset might overcome these issues.%
\footnote{Instead of feature engineering, model selection or neural architecture search requires domain knowledge to solve machine learning problems with representation learning; however, this can be automated by using machine learning thanks to recent advances of AutoML~\citep{hutter2019automated} and NAS~\citep{Elsken2018JMLR}, respectively.}
We believe that this is why representation learning has played an important role in recent machine learning.
Generally, an extracted feature representation forms a real-valued vector by using representation learning algorithms.
Thanks to this property, we seamlessly introduce a feature representation by replacing feature engineering with a learned feature representation in an existing machine learning system or data analysis platform pipeline.
Learned representations tend to be called ``distributed representations''~\citep{Hinton1986BookChapter} or ``(feature) embeddings''.
Unlike local representation, elements in distributed representation do not have a clear meaning.
For example, if we learn a feature representation of word ``game'', the distributed representation of ``game'' is like
\begin{align}
  \hbf(\xbf_{\text{game}}) =
  [\overbrace
  {0.06, 0.01, -0.49, 0.27, 0.38, \ldots}
  ^{d}].
  \label{eq:word-rep-vec}
\end{align}
Of course, we can make representations more interpretable by introducing a regularization such as the sparsity~\citep{Ranzato2006NeurIPS} or disentanglement~\citep{Higgins2018arXiv}.
Learned representations require less capacity to represent many patterns than local representations.
Note that we believe that \citet{Hinton1986BookChapter} conceptually introduced distributed representations rather than just real-value vectors extracted using a machine learning model as in recent machine learning papers.
However, following the terminology in the recent machine learning community,
we discuss feature vectors extracted by feature extractors in this review.

\section{Evaluation Methods of Representation Learning}
We now organize the evaluation methods of representation learning algorithms.
We do not discuss general evaluation metrics for machine learning algorithms, such as computing efficiency that is not simple; a single efficiency metric is misleading in deep neural networks as discussed by~\citet{Dehghani2022ICLR}.
For all evaluation perspectives except for ``Representation Learning as an Auxiliary Task'' described in~\cref{sec:auxiliary},
suppose that we have $R$ pre-trained representation learning models, $\{ \widehat{\hbf}_r \}_{r=1}^R$ such as different representation learning algorithms or different hyperparameter configurations.
Given $R$ pre-trained models, we would like to determine the best one.

\subsection{Representation Learning for Pre-training}
\label{sec:pre-training}

Since representations play an important role in solving machine learning problems, as explained in~\cref{sec:introduction}, we expect that extracted representations by a representation learning algorithm generalize to unseen machine learning tasks: downstream tasks, such as classification.
Motivated by this expectation, the most common evaluation method is how learned representations help solve downstream tasks.
In this sense, we consider representation learning the pre-training~\citep{Hinton2006NC} of the feature extractor of downstream tasks.
Note that recently we no longer use layer-wise greedy pre-training as proposed by \citet{Hinton2006NC} to train deep neural networks, thanks to the progress in learning techniques of deep learning.%
\footnote{Unsupervised pre-training~\citep{Hinton2006NC} is a historically important two-stage training algorithm.
Unsupervised pre-training tackles the difficulty of training deep brief nets (DBNs) that are restricted Boltzmann machine-based generative models.
\citet{Hinton2006NC} proposed a training procedure for deep belief networks:
(i) a greedy layer-wise unsupervised pre-training algorithm to initialize the weights of DBNs and
(ii) fine-tuning of pre-trained weights.
Similarly, \citet{Vincent2008ICML} proposed a greedy pre-training algorithm by using a stacking denoising auto-encoder instead of deep brief nets.
Thanks to the pre-training, optimization of deep models converges a better local optimum than random initialization~\citep{Hinton2006NC,Vincent2008ICML}.}

As a running example, we suppose a classification problem as a downstream task.
Downstream dataset is denoted $\Dcal_D = \{ (\xbf_i, y_i) \}_{i=1}^{N_D}$, where $N_D$ is the number of samples and $y_i \in \Ycal$ is a class label.
Downstream dataset $\Dcal_D$ can be the same as the dataset of representation learning to obtain $\widehat{\hbf}$.
Suppose that pre-trained model $\widehat{\hbf}$ is used in the model of the downstream task denoted by $\hbf_D$.
For example, both $\widehat{\hbf}$ and ${\hbf}_D$ are the same neural networks to extract feature representations from $\xbf$ or pre-trained word embeddings with the same dimensionalities.
We use the pre-trained parameters of $\widehat{\hbf}$ as initialization values of the parameters of $\hbf_D$.
To solve the downstream task, we require an additional function $\gbf_D$ that maps feature space to label space: $\Rbb^d \rightarrow \Rbb^{|\Ycal|}$ since $\widehat{\hbf}$ is designed to extract feature representations, not to solve the downstream task.
For example, in recent self-supervised learning on computer vision~\citep{Bachman2019NeurIPS,He2020CVPR}, ResNet-50~\citep{He2016CVPR} was used as $\widehat{\hbf}$ and $\hbf_D$ to extract feature representations, and a linear classifier was implemented as $\gbf_D$.
We tend to implement $\gbf_D$ as a simple function such as logistic regression, support vector machines, or shallow neural networks.
This is because such a simple $\gbf_D$ is enough to solve the downstream task if extracted representations already capture discriminative features~\citep{Bengio2013IEEE}.
\cref{table:pre-training-table} categorizes studies that condcut pre-training-based experiments.

\begin{table*}
  \centering
  {\scriptsize
  \begin{tabular}[t]{ccccccccc}
    \toprule
    References & Domain & Upstream task & Downstream task & \multicolumn{3}{c}{Frozen} & Fine-tuning & Theoretical analysis \\
    \cmidrule{5-7} &&&& Linear & & Nonlinear & & \\
    \midrule

    \citet{He2016CVPR} & CV      & Sup. & Trans.       &            & & & \checkmark \\
    \citet{He2021ICLR} & NLP     & Uns. & Trans.       &            & & & \checkmark \\
    \citet{Wang2019NeurIPSa} & NLP & Uns. & Trans.       & \checkmark & & & \checkmark \\
    \citet{Mikolov2013ICLR} & NLP& Uns. & Trans.       & \checkmark & & &  \\
    \citet{Mikolov2013NeurIPS} & NLP & Uns. & Trans.       & \checkmark & & &  \\
    \citet{Perozzi2014KDD} & Graph & Uns. & Sup.        & \checkmark & & &  \\
    \citet{Tang2015WWW} & Graph & Uns. & Sup.        & \checkmark & & &  \\
    \citet{Devlin2019NAACL} & NLP & Uns. & Trans.       & & & \checkmark & \checkmark \\
    \citet{Schnabel2015EMNLP} & NLP & Uns. & Trans. & \checkmark & & \checkmark & \\
    \citet{Radford2021ICML} & CV   & Weak & Trans.       & \checkmark & & & \checkmark \\
    \citet{Oord2018arXiv} & CV, NLP, Speech & Uns. & Sup., Trans. & \checkmark & & & & \\
    \citet{Donahue2014ICML} & CV & Sup. & Trans. & \checkmark & & & \\
    \citet{Yosinski2014NeurIPS} & CV & Sup. & Sup., Trans. & & & \checkmark & \checkmark \\
    \citet{Hjelm2019ICLR} & CV & Uns. & Sup.  & \checkmark & & \checkmark & \checkmark & \\
    \citet{Bachman2019NeurIPS} & CV & Uns. & Sup., Trans. & \checkmark & &   \checkmark \\
    \citet{Baevski2021NeurIPS}& Speech & Uns. & Trans. & \checkmark & & \checkmark &  \\
    \citet{Abu-El-Haija2016arXiv} & CV & Sup. & Sup., Trans.  & \checkmark & & \checkmark & \checkmark \\
    \citet{Conneau2017EMNLP} & NLP & Sup. & Sup., Trans. & \checkmark & & & \\
    \citet{Bowman2015EMNLP}   & NLP & Sup. & Trans. & & & & \checkmark \\
    \citet{McCann2017NeurIPS} & NLP & Sup. & Trans. & & & \checkmark & \\
    \citet{Long2015CVPR} & CV & Sup. & Trans. & & & & \checkmark \\
    \citet{Girshick2014CVPR} & CV & Sup. & Trans. & \checkmark & & & \checkmark \\
    \citet{Girshick2015ICCV} & CV & Sup. & Trans. & & & & \checkmark \\
    \citet{Ren2015NeurIPS} & CV & Sup. & Trans. & & & \checkmark & \checkmark \\
    \citet{Sun2017ICCV} & CV & Sup. & Trans. & & & \checkmark & \checkmark \\
    \citet{Mahajan2018ECCV} & CV & Weak & Trans. & \checkmark & & & \checkmark \\
    \citet{Kolesnikov2020ECCV} & CV & Sup. & Sup., Trans. & & & & \checkmark \\
    \citet{Kataoka2020ACCV} & CV & Sup. & Trans. & & & & \checkmark \\
    \citet{Kornblith2019CVPR} & CV & Sup. & Trans. & \checkmark & & & \checkmark \\
    \citet{Abnar2022ICLR} & CV & Sup. & Trans. & \checkmark & & & \checkmark \\
    \citet{Ericsson2021CVPR} & CV & Sup., Uns.  & Sup., Trans. & \checkmark & & & \checkmark \\
    \citet{Mikolov2018LREC} & NLP & Uns. & Trans. & \checkmark & & & \\
    \citet{Du2021arXiv} & NLP & Uns. & Trans. & \checkmark & & & \\
    \citet{He2020CVPR} & CV & Sup., Uns. & Sup., Trans. & \checkmark & & & \checkmark \\
    \citet{Caron2020NeurIPS} & CV & Sup., Uns. & Sup., Trans. & \checkmark & & & \checkmark \\
    \citet{Goyal2021arXiv} & CV & Uns. & Trans. & \checkmark & & & \checkmark \\
    \citet{Zhai2020arXiv} & CV & Semi, Sup., Uns. & Trans. & \checkmark & & & \checkmark \\
    \citet{Hernandez2021arXiv} & NLP & Uns. & Trans. & & & & \checkmark \\
    \citet{Kotar2021ICCV} & CV & Sup., Uns. & Sup., Trans. & \checkmark & & \checkmark & \\
    \citet{Sariyildiz2021ICCV} & CV & Sup., Uns. & Trans. & \checkmark & & & \\
    \citet{Mitrovic2021ICLR} &  CV & Uns. & Sup., Trans. & \checkmark & & & & \checkmark \\
    \citet{Kolesnikov2019CVPR} & CV & Uns. & Sup. & \checkmark & & \checkmark & \\
    \citet{Ranzato2006NeurIPS} & CV & Uns. & Sup. & & & & \checkmark \\
    \citet{Hinton2006NC} & CV & Uns. & Sup. & & & & \checkmark \\
    \citet{Vincent2008ICML} & CV & Uns. & Sup. & & & & \checkmark \\
    \citet{Henaff2020ICML} & CV & Sup. & Sup., Trans. & \checkmark & & & \checkmark \\
    \citet{Baroni2014ACL} & NLP & Uns. & Trans. & \checkmark & & & \\
    \citet{Levy2014NeurIPS} & NLP   & Uns. & Trans. & \checkmark & & & &  \\
    \citet{Arora2019ICML} & CV, NLP & Uns. & Sup. & \checkmark & & & & \checkmark\\
    \citet{Zhai2019ICCV} & CV & Semi & Trans. & \checkmark & & & \\
    \citet{Saunshi2021ICLR} & NLP & Uns. & Trans. & \checkmark & & & \checkmark & \checkmark \\
    \citet{Zhao2021ICLR} & CV & Sup., Uns. & Sup., Trans. & \checkmark & & & \checkmark \\
    \citet{Nozawa2021NeurIPS} & CV & Uns. & Sup. & \checkmark & & & & \checkmark\\
    \citet{Islam2021ICCV} &  CV & Sup., Uns. & Sup., Trans. & \checkmark & & & \checkmark \\
    \citet{Tamkin2021NeurIPSData} & CV, NLP, Sensor, Speech & Uns. & Sup., Trans. & \checkmark & & & \checkmark \\
    \citet{Liu2022ICLR} &    CV  & Sup., Uns. & Sup., Trans. & \checkmark & & & \checkmark & \checkmark\\
    \bottomrule
  \end{tabular}
  \caption{
    Categorized studies into pre-training evaluation.
    In the column of the upstream task, ``Uns'' stands for unsupervised representation learning,
  ``Sup.'' stands for supervised representation learning, and ``Semi'' stands for semi-supervised representation learning.
    In the column of the downstream task, ``Sup.'' means that the authors use the same dataset as the upstream task to evaluate the feature extractor, and ``Trans.'' means that the authors use a different dataset from the upstream task.
    Note that an identity predictor is also categorized as ``Linear'' in the ``Frozen'' column.
    We add a mark in the theoretical analysis column if the study performed an analysis in this evaluation setting.
    }
  \label{table:pre-training-table}
  }
\end{table*}

\begin{table}[h!]
  \ContinuedFloat  %
  \centering
  {\scriptsize
  \begin{tabular}[t]{ccccccccc}
    \toprule
    References & Domain & Upstream task & Downstream task & \multicolumn{3}{c}{Frozen} & Fine-tuning & Theoretical analysis \\
    \cmidrule{5-7} &&&& Linear & & Nonlinear &  \\
    \midrule
    \citet{Erhan2010JMLR} & CV & Uns. & Sup. & & & & \checkmark \\
    \citet{Kong2020ICLR} & CV & Uns. & Trans. & & & & \checkmark & \checkmark \\
    \citet{Wieting2019ICLR} & NLP & Uns. & Trans. & \checkmark & & \checkmark &  \\
    \citet{Chen2020ICML} & CV & Uns. & Sup., Trans. & \checkmark & & & \checkmark \\
    \citet{Grill2020NeurIPS} & CV & Uns. & Sup., Trans. & \checkmark & & & \checkmark \\
    \citet{He2021arXiv} & CV & Sup., Uns. & Sup., Trans. & \checkmark & & & \checkmark \\
    \citet{Newell2020CVPR} & CV & Uns. & Sup. & \checkmark & & & \checkmark \\
    \citet{Misra2020CVPR} & CV & Sup., Uns. & Sup., Trans. & \checkmark & & & \checkmark \\
    \citet{Goyal2019ICCV} & CV & Sup., Uns. & Sup., Trans. & \checkmark & & & \checkmark \\
    \citet{Brown2020NeurIPS} & NLP & Uns. & Trans. & \checkmark & & &  \\
    \citet{Musgrave2020ECCV} & CV & Sup. & Trans. & \checkmark & & & \checkmark \\
    \citet{Chen2019ICLR} & CV & Sup. & Trans. & \checkmark & & & \\
    \citet{He2019CVPR} & CV & Sup. & Trans. & & & & \checkmark \\
    \citet{Oliver2018NeurIPS} & CV & Sup. & Trans. & & & & \checkmark \\
    \citet{Gidaris2018ICLR} & CV & Uns. & Sup., Trans. & \checkmark & & \checkmark & \checkmark \\
    \citet{Peters2018NAACL}& NLP & Uns. & Trans. & \checkmark & & \checkmark & \checkmark \\
    \citet{Tian2020ECCV_a} & CV  & Sup., Uns. & Trans. & \checkmark & & \checkmark & \\
    \citet{Medina2020arXiv}& CV  & Sup., Uns. & Trans. & \checkmark & & & \\
    \citet{Nozawa2020UAI}  & CV, Time-series & Sup., Uns. & Sup. & \checkmark & & &  & \checkmark\\
    \citet{Chuang2020NeurIPS}& CV & Uns. & Sup. & \checkmark & & &  & \checkmark \\
    \citet{Lee2021NeurIPS} & CV, NLP & Uns. & Sup. & \checkmark & & &  & \checkmark\\
    \citet{Wei2021NeurIPS} & NLP & Uns. & Sup. & \checkmark & & &  & \checkmark\\
    \citet{Wei2021ICLR}      & CV, NLP & Sup. & Trans. & \checkmark & & &  & \checkmark\\
    \citet{HaoChen2021NeurIPS} & CV & Uns. & Sup. & \checkmark & & &  & \checkmark\\
    \citet{Wang2022ICLR}     & CV & Uns. & Sup. & \checkmark & & &  & \checkmark\\
    \citet{Bansal2021ICLR}   & CV & Uns. & Sup. & \checkmark & & \checkmark & & \checkmark\\
    \citet{Du2021ICLR}       & N/A & Sup. & Trans. & \checkmark & & & & \checkmark\\
    \citet{McNamara2017ICML} & CV, NLP & Sup. & Sup., Trans. & \checkmark & & & \checkmark & \checkmark\\
    \citet{Tosh2021JMLR}     & NLP & Uns. & Sup. & \checkmark & & & & \checkmark\\
    \citet{Tschannen2020ICLR}& CV & Uns. & Sup. & \checkmark & & & \\
    \citet{Hashimoto2016TACL}& NLP & Uns. & Trans. & \checkmark & & &  & \\
    \citet{Grover2016KDD}    & Graph & Uns. & Sup. & \checkmark & & & \\
    \citet{Levy2015TACL}     & NLP & Uns. & Trans. & \checkmark & & & \\
    \citet{Allen2019NeurIPS} & NLP & Uns. & Trans. & \checkmark & & &  & \\
    \citet{Qiu2018WSDM}      & Graph & Uns. & Sup. & \checkmark & & &  & \\
    \citet{Arora2016TACL}    & NLP & Uns. & Trans. & \checkmark & & &  & \\
    \citet{Li2021NeurIPS} & CV & Uns. & Sup., Trans. & \checkmark & & & \checkmark  & \\
    \citet{Tian2020NeurIPS} & CV & Uns. & Sup., Trans. & \checkmark & & & \checkmark \\
    \citet{Wang2020ICML} & CV, NLP & Uns. & Sup., Trans. & \checkmark & & \checkmark & \checkmark & \\
    \citet{Dubois2020NeurIPS} & CV & Sup. & Sup. & \checkmark & & & & \checkmark \\
    \citet{Gidaris2019ICCV}   & CV & Semi., Sup., Uns. & Trans. & \checkmark & & & \\
    \citet{Dosovitskiy2014NeurIPS}& CV & Uns. & Trans. & \checkmark & & & \\
    \citet{Doersch2015ICCV}& CV & Uns. & Sup., Trans. & & & & \checkmark \\
    \citet{Noroozi2016ECCV}& CV & Uns. & Sup., Trans. & & & & \checkmark \\
    \citet{Kiros2015NeurIPS}& NLP & Uns. & Trans. & \checkmark & & & \\
    \citet{Dai2015NeurIPS} & CV, NLP & Uns. & Sup., Trans. &  & & & \checkmark \\
    \citet{Tosh2021ALT} & N/A & Uns. & Sup. & \checkmark & & & \checkmark & \checkmark
    \\
    \citet{Kumar2022ICLR} & CV & Sup. & Trans. & \checkmark  & & & \checkmark & \checkmark \\
    \citet{Pennington2014EMNLP} & NLP & Uns. & Trans. & \checkmark  & & \checkmark &  & \\
    \citet{Chen2021CVPR} & CV & Uns. & Sup., Trans. & \checkmark  & & & \checkmark & \\
    \citet{Jing2022ICLR}  & CV & Uns. & Sup. & \checkmark  & & & & \\
    \citet{Wen2021ICML} & N/A & Uns. & Sup. & \checkmark  & & & & \\
    \citet{Tian2021ICML} & CV & Uns. & Sup. & \checkmark  & & & & \\
    \bottomrule
  \end{tabular}
  }
  \caption{
    (Continued)
    Categorized studies into pre-training evaluation.
    In the column of the upstream task, ``Uns'' stands for unsupervised representation learning,
  ``Sup.'' stands for supervised representation learning, and ``Semi'' stands for semi-supervised representation learning.
    In the column of the downstream task, ``Sup.'' means that the authors use the same dataset as the upstream task to evaluate the feature extractor, and ``Trans.'' means that the authors use a different dataset from the upstream task.
    Note that an identity predictor is also categorized as ``Linear'' in the ``Frozen'' column.
    We add a mark in the theoretical analysis column if the study performed an analysis in this evaluation setting.
  }
\end{table}

\subsubsection{Experimental Procedure}
\label{sec:pre-training-experimental-procedure}

\begin{figure}[th]
  \centering
  \resizebox{0.7\textwidth}{!}{%
\begin{tikzpicture}[shorten >=1pt,->,draw=black!50, node distance=\layersep]
    \tikzstyle{every pin edge}=[<-,shorten <=1pt]
    \tikzstyle{neuron}=[circle,fill=black!25,minimum size=15pt,inner sep=0pt]
    \tikzstyle{dummy}=[circle,fill=white!25,minimum size=15pt,inner sep=0pt]
    \tikzstyle{input neuron}=[neuron]
    \tikzstyle{output neuron}=[neuron]
    \tikzstyle{hidden neuron}=[neuron]
    \tikzstyle{annot} = [text width=5em, text centered]

    \foreach \name / \y in {1,...,4}
        \node[input neuron, pin=left:$x_{\y}$] (I-\name) at (-47mm,- \y cm) {};

    \foreach \name / \y in {1,...,5}
        \path[yshift=0.5cm]
            node[hidden neuron] (H-\name) at (\layersep-47mm,-\y cm) {};

    \foreach \name / \y in {1,...,5}
        \path[yshift=0.5cm]
            node[hidden neuron] (HH-\name) at (2 * \layersep-47mm,-\y cm) {};

    \foreach \name / \y in {1,...,4}
            \node[output neuron] (O-\name) at (3 * \layersep-47mm,-\y cm) {};

    \foreach \source in {1,...,4}
        \foreach \dest in {1,...,5}
            \path[color=red] (I-\source) edge (H-\dest);

    \foreach \source in {1,...,5}
        \foreach \dest in {1,...,5}
            \path[color=red] (H-\source) edge (HH-\dest);

    \foreach \source in {1,...,5}
        \foreach \dest in {1,...,4}
        \path[color=red] (HH-\source) edge (O-\dest);

    \draw [
        decorate,
        -,
        thick,
        decoration={
            brace,
            amplitude=3pt,
            mirror,
            raise=0.5cm,
        }
    ] (HH-5.west -| I-4.east) -- (HH-5)
    node [pos=0.5,anchor=north,yshift=-0.55cm] {\footnotesize Feature extractor $\hbf$};

    \draw [
        decorate,
        -,
        thick,
        decoration={
            brace,
            amplitude=3pt,
            mirror,
            raise=0.5cm,
        }
    ]
    (HH-5) -- (HH-5.east -| O-1.west)
    node [pos=0.5,anchor=north,yshift=-0.55cm] {\footnotesize Task specific head $\gbf$};

    \node[dummy] (dummy1) at (1.5 * \layersep-48mm,-6 cm) {};

    \node[annot] (hl) at (1.5 * \layersep-48mm, 0.3cm) {Pretraining};

    \begin{scope}[shift={(-80mm,-74mm)}]
        \foreach \name / \y in {1,...,4}
        \node[input neuron, pin=left:$x_{\y}$] (I-\name) at (0,-\y cm) {};

        \foreach \name / \y in {1,...,5}
            \path[yshift=0.5cm]
                node[hidden neuron] (H-\name) at (\layersep,-\y cm) {};

        \foreach \name / \y in {1,...,5}
            \path[yshift=0.5cm]
                node[hidden neuron] (HH-\name) at (2 * \layersep,-\y cm) {};

        \node[output neuron] (O) at (3 * \layersep,-2.5 cm) {};

        \foreach \source in {1,...,4}
            \foreach \dest in {1,...,5}
                \path[color=blue] (I-\source) edge (H-\dest);

        \foreach \source in {1,...,5}
            \foreach \dest in {1,...,5}
            \path[color=blue] (H-\source) edge (HH-\dest);

        \foreach \source in {1,...,5}
            \path[color=red] (HH-\source) edge (O);

        \draw [
            decorate,
            -,
            thick,
            decoration={
                brace,
                amplitude=3pt,
                mirror,
                raise=0.5cm,
            }
        ] (HH-5.west -| I-4.east) -- (HH-5)
        node [pos=0.5,anchor=north,yshift=-0.55cm] {\footnotesize Fixed $\widehat{\hbf}$};

        \draw [
            decorate,
            -,
            thick,
            decoration={
                brace,
                amplitude=3pt,
                mirror,
                raise=0.5cm,
            }
        ]
        (HH-5) -- (HH-5.east -| O.west)
        node [pos=0.5,anchor=north,yshift=-0.55cm] {\footnotesize Task specific head $\gbf_D$};

        \node[dummy] (dummy2) at (1.5 * \layersep,0.5 cm) {};

        \node[annot] (hl) at (1.5 * \layersep, 0.3cm) {Frozen};
    \end{scope}

    \begin{scope}[shift={(5mm,-74mm)}]
        \foreach \name / \y in {1,...,4}
        \node[input neuron, pin=left:$x_{\y}$] (I-\name) at (0,-\y cm) {};

        \foreach \name / \y in {1,...,5}
            \path[yshift=0.5cm]
                node[hidden neuron] (H-\name) at (\layersep,-\y cm) {};

        \foreach \name / \y in {1,...,5}
            \path[yshift=0.5cm]
                node[hidden neuron] (HH-\name) at (2 * \layersep,-\y cm) {};

        \node[output neuron] (O) at (3 * \layersep,-2.5 cm) {};

        \foreach \source in {1,...,4}
            \foreach \dest in {1,...,5}
                \path[color=violet] (I-\source) edge (H-\dest);

        \foreach \source in {1,...,5}
            \foreach \dest in {1,...,5}
                \path[color=violet] (H-\source) edge (HH-\dest);

        \foreach \source in {1,...,5}
            \path[color=red] (HH-\source) edge (O);

        \draw [
            decorate,
            -,
            thick,
            decoration={
                brace,
                amplitude=3pt,
                mirror,
                raise=0.5cm,
            }
        ] (HH-5.west -| I-4.east) -- (HH-5)
        node [pos=0.5,anchor=north,yshift=-0.55cm] {\footnotesize Initialized with $\widehat{\hbf}$};

        \draw [
            decorate,
            -,
            thick,
            decoration={
                brace,
                amplitude=3pt,
                mirror,
                raise=0.5cm,
            }
        ]
        (HH-5) -- (HH-5.east -| O.west)
        node [pos=0.5,anchor=north,yshift=-0.55cm] {\footnotesize Task specific head $\gbf_D$};

        \node[dummy] (dummy3) at (0.35 * \layersep,0.5 cm) {};

        \node[annot] (hl) at (\layersep, 0.3cm) {Fine-tuning};
    \end{scope}

\draw[black, line width=0.7pt] (dummy1.north) -- (dummy2);
\draw[black, line width=0.7pt] (dummy1.north) -- (dummy3);

\end{tikzpicture}

}
  \caption{
    Overview of pre-training approach.
    Edge represents the weights of neural networks.
    Red-colored weights in neural networks are initialized randomly.
    In frozen protocol, blue-colored weights are initialized by using pre-trained feature extractor $\widehat{\hbf}$ and task-specific head $\gbf_D$ is initialized randomly.
    During optimization, we fix blue-colored weights and train only $\gbf_D$.
    In fine-tuning protocol, purple-colored weights in a neural network are also initialized by using $\widehat{\hbf}$.
    During optimization, we train purple-colored weights and $\gbf_D$.
  }
  \label{fig:overview-pre-training}
\end{figure}
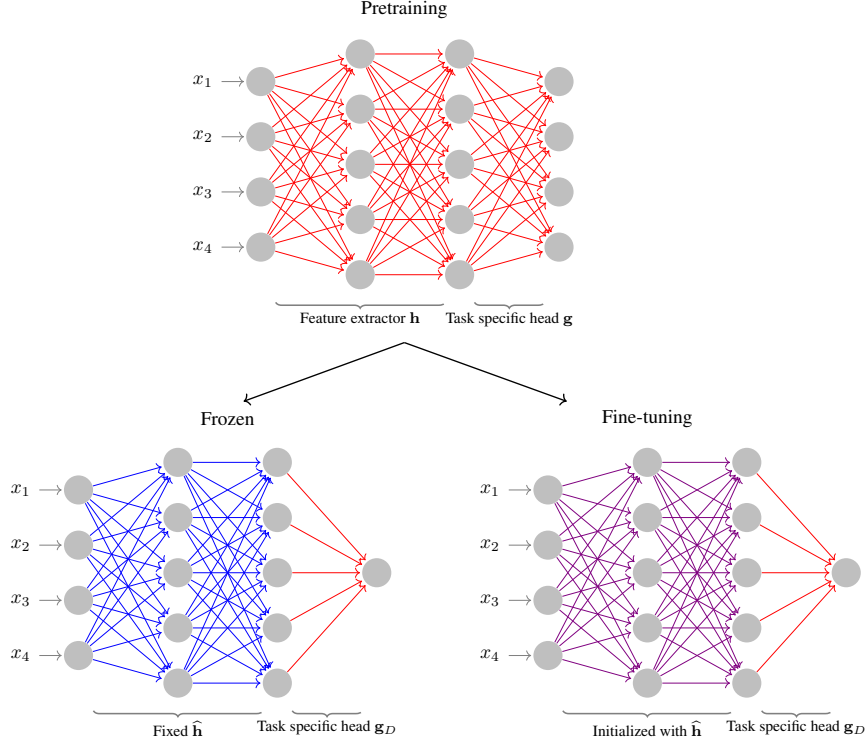

Given $R$ pre-trained feature extractors $ \{ \widehat{\hbf}_r \}_{r=1}^R$ and downstream dataset $\Dcal_D$,
we compare the extractors by using the evaluation metric of the downstream task.
This is equivalent to treating the feature extractor as a hyper-parameter in the model of the downstream task.
The evaluation procedures are as follows:
\begin{enumerate}
  \item Train $\hbf_D, \gbf_D$ on downstream dataset $\Dcal_D$ with pre-trained feature extractor $\widehat{\hbf}_r$ for each $r$.
  \item Compare evaluation metric values of the downstream task, such as validation accuracy.
\end{enumerate}
In the first step, there are two common protocols to evaluate feature extractors: ``frozen'' and ``fine-tuning''.
\Cref{fig:overview-pre-training} illustrates the overview of these evaluation protocols.

\paragraph{Frozen protocol}
This evaluation protocol has been quite common in recent representation learning experiments.
Since we expect that $\widehat{\hbf}$ can extract discriminative features for the downstream task, we do not update $\hbf_D$ initialized by $\widehat{\hbf}$ during the training of the downstream task.
Training $\gbf_D$ requires less computing budget and converges faster than the training of $\hbf_D$ and $\gbf_D$ from scratch on the downstream dataset.
Formally, we solve the following problem:
\begin{align}
  \min_{\gbf_D} \widehat{L}_D(\widehat{\hbf}, \gbf_D),
  \label{eq:frozen-loss}
\end{align}
where $\widehat{L}_D$ is an empirical risk on $\Dcal_D$ such as softmax cross-entropy loss \eqref{eq:supervised-loss} for classification.
The standard choice of $\gbf_D$ is a linear classifier~\citep{Donahue2014ICML,He2020CVPR} or non-parametric method, such as $k$-nearest neighbors.
When we use a linear classifier as $\gbf_D$, the evaluation protocol is also called ``linear probing''.
To attain a further performance gain with additional computing cost, we implement $\gbf_D$ as a nonlinear model, for example, shallow neural networks with a nonlinear activation function~\citep{Bachman2019NeurIPS}, Long Short-Term Memory, or echo state networks~\citep{Wieting2019ICLR}.

Especially in NLP, ``Prompt''~\citep{Liu2021arXiv} is a new method to solve a downstream task given a language model or masked language model~\citep{Devlin2019NAACL}.
Depending on the prompt's formulation, we do not require any training of model parameters on $D_D$.
Instead, we need to design a text interface, called prompt, such that the model fills a blank part in the prompt to solve the downstream task.
For example, for binary news topic classification, ``science'' vs ``art'',
the prompt is
\begin{align*}
  \text{\texttt{[NEWS] This article is about [CLASS]}}.
\end{align*}
For inference, the language model takes input sentence $\xbf$ as \texttt{[NEWS]} and a word in pre-defined words to represent a topic, such as \texttt{science} or \texttt{art}, as \texttt{[CLASS]}.
We obtain a predicted class by selecting a word that maximizes the probability of the output of the language model in the pre-defined words.

\paragraph{Fine-tuning protocol}
To achieve further performance gain of the downstream task or fill the gap between representation learning and downstream tasks, we train both $\hbf_D$ initialized by $\widehat{\hbf}$ and $\gbf_D$ as a single model on the downstream task.
This procedure is called ``fine-tuning''.
Formally, we solve the following problem:
\begin{align}
  \min_{\hbf_D, \gbf_D} \widehat{L}_D(\hbf_D, \gbf_D),
  \text{ where } \hbf_D \text{ is initialized by } \widehat{\hbf}.
  \label{eq:fine-tuning-loss}
\end{align}
We might update $\hbf_D$ with a smaller learning rate in gradient descent-based optimization than randomly initialized weights~\citep{Girshick2014CVPR}.
This is because we expect that $\hbf_D$ with pre-trained weights has already been able to extract useful feature representations for the downstream task.
If we set an optimizer's inappropriate hyper-parameters, such as too large a learning rate or too many iterations, $\hbf_D$ is likely to forget the pre-trained weights.
As a result, the model overfits $\Dcal_D$.
To avoid this explicitly, we can use $\widehat{\hbf}$ as a regularizer as reviewed in~\cref{sec:regularization}.
As another technique, the combination of linear probing and fine-tuning protocol is also proposed, especially for out-of-distribution because the feature extractor's performance on data sampled from out-of-distribution degrades after fine-tuning~\citep{Kumar2022ICLR}.
Even though the fine-tuning protocol requires more computing budget than the frozen protocol,
it empirically performs better than the frozen protocol~\citep{Zhai2019ICCV,Chen2020ICML,Grill2020NeurIPS}.
Notably, we suggest tuning hyper-parameters among these two protocols independently because the optimal hyper-parameters for two protocols are different~\citep{Zhai2020arXiv,He2021arXiv} or even uncorrelated~\citep{Newell2020CVPR}.

\paragraph{Efficiency}
Complementary to the two evaluation protocols, varying the size of the downstream dataset and computing budget is concerned in representation learning experiments~\citep{Henaff2020ICML,Misra2020CVPR}.
Intuitively, if feature extractor $\widehat{\hbf}$ captures discriminative feature representations, training $\hbf_D$ and $\gbf_D$ requires fewer labeled data or less computing budget, i.e., fewer epochs in a gradient descent algorithm, than the same model with random initialization to achieve similar generalization performance~\citep{Erhan2010JMLR}.

\subsubsection{Discussion}
\label{sec:discussion-pre-training}

\paragraph{Fair Comparison}
We need to pay attention to the size of representation learning data and feature extractor $\widehat{\hbf}$ to compare different representation learning algorithms.
For deep neural network-based representation learning algorithms, there exists a positive correlation between the model size of $\widehat{\hbf}$ and downstream performance,
for example, \citet{Bachman2019NeurIPS,He2020CVPR,Goyal2021arXiv,Kolesnikov2019CVPR,Goyal2019ICCV} for vision and~\citet{Devlin2019NAACL,Brown2020NeurIPS} for language.
Enlarging the size of data makes this tendency stronger~\citep{Kolesnikov2020ECCV}.
Suppose we propose a novel representation learning algorithm to improve the state-of-the-art performance on downstream tasks.
In this case, we should use the same architecture and dataset to disentangle the factors of performance gain.
\citet{Musgrave2020ECCV,Chen2019ICLR} discuss
the same problem in metric learning and few-shot learning, respectively.
If we propose a representation learning algorithm as the pre-training of a specific model in a downstream task rather than unknown downstream tasks,
we should treat representation learning as a part of downstream optimization by following the discussion~\citep{He2019CVPR} for a fair comparison, especially in terms of computing cost.
We also highly recommend following suitable suggestions for representation learning experiments by \citet{Oliver2018NeurIPS}.

\paragraph{Overfitting benchmark datasets}
One concern of this evaluation procedure, especially on a single dataset, is the overfitting to a benchmark downstream task.
For example, recent representation models are trained on the ImageNet-1K dataset~\citep{Deng2009CVPR} that is class balanced and object-centric classification dataset.
We do not know that proposed representation learning algorithms only for ImageNet work well on largely different datasets such as cartoon%
\footnote{\url{https://google.github.io/cartoonset/}}
or medical images.
In fact, the recent self-supervised learning in the vision domain heavily depends on data-augmentation techniques that might not be able to be applied to another dataset.
For example, we cannot apply the rotation-based representation learning algorithm~\citep{Gidaris2018ICLR} to the MNIST classification task because we should distinguish the class ``6'' and the class ``9'' with $180^{\circ}$ rotation.

\paragraph{Best representations in the layers of neural networks}
For deep neural network-based models, we have multiple candidates of $\widehat{\hbf}$ depending on which sub-network we select as a feature extractor.
The optimal feature extractor among the layers depends on the downstream task.
Concretely, the representations extracted by using until the last layer tend to specialize in the representation task~\citep{Yosinski2014NeurIPS}.
As a result, such $\widehat{\hbf}$ performs poorly as a feature extractor on the downstream task, especially without fine-tuning.
Indeed, removing the last few layers from a neural network is a technique to improve the downstream performance in practice~\citep{Rogers2020TACL,Donahue2014ICML,Hjelm2019ICLR,Bachman2019NeurIPS,Baevski2021NeurIPS,Girshick2014CVPR,Goyal2019ICCV}.
We recommend trying different intermediate representations as a hyperparameter of the downstream task, especially in the frozen protocol.
Empirically, we can use a combination of multiple intermediate representations~\citep{Peters2018NAACL,Pennington2014EMNLP} because extracted feature vectors capture different information depending on the layer.

\paragraph{Relation to other machine learning settings}
The described experimental protocols are similar to transfer learning settings~\citep{Pan2010IEEE}.
The frozen protocol and fine-tuning protocol are similar to ``feature-representation-transfer'' and ``parameter transfer'', respectively.
In transfer learning terminology, we train feature extractor $\hbf$ on a source task,
and then we transfer pre-trained $\widehat{\hbf}$ to a target task.
In addition, few-shot learning~\citep{Wang2020ACM}, where the labeled dataset contains few labeled samples per class, can be used to evaluate representation learning~\citep{Brown2020NeurIPS,Tian2020ECCV_a,Medina2020arXiv}.

Another setting is semi-supervised learning~\citep{Chapelle2006Book}, where we train a predictor from many unlabeled data and a few labeled data.
Since unsupervised representation learning does not require a labeled dataset,
we train $\hbf$ on the unlabeled data, then train $\hbf_{\mathrm{D}}$ and $\gbf_D$ with $\widehat{\hbf}$ on the labeled data~\citep{Zhai2019ICCV}.
We will discuss other representation learning-based approaches for the semi-supervised learning scenario in the other evaluation perspectives described in~\cref{sec:regularization,sec:auxiliary}.

\paragraph{Benefits for optimization}
As described above, pre-trained weights of representation learning model $\widehat{\hbf}$ behave as the initialization of downstream task's model $\hbf_D$.
Since an initialization method is a key factor to improve performance in the gradient-based optimization of deep neural nets~\citep{Sutskever2013ICML}, pre-trained models help the optimization of the downstream task.
Concretely, we can compare representation learning algorithms in terms of stability~\citep{Erhan2010JMLR}.
Suppose the pre-trained feature extractor $\widehat{\hbf}$ is the good initialization of $\hbf_D$.
In that case, the variance of optimum among multiple runs with different random seeds is smaller than random initialization, which means the pre-trained $\widehat{\hbf}$ is robust initialization to the randomness of the training for the downstream task.

\subsection{Representation Learning for Regularization}
\label{sec:regularization}

\begin{table*}
  \centering
  \begin{tabular}{cccc}
    \toprule
    Reference & Main task & The task of $\widehat{\hbf}$ & Type of  $\Omega$ \\
    \midrule
    \citet{Romero2015ICLR} & Supervised classification & Same classification & \cref{eq:single-regularization} \\
    \citet{Tian2020ECCV_a} & Transfer learning & Classification on source task & \cref{eq:single-regularization} \\
    \citet{McNamara2017ICML} & Transfer learning & Classification on source task & \cref{eq:whole-regularization} \\
    \citet{Li2018ICML} & Transfer learning & Classification on source task & \cref{eq:whole-regularization} \\
    \bottomrule
  \end{tabular}
  \caption{
    The categorized experimental settings in \cref{sec:regularization}.
  }
  \label{table:regularizations}
\end{table*}

Even though we fine-tune the weights of $\hbf_D$ initialized with pre-trained $\widehat{\hbf}$ as described in~\cref{sec:pre-training},
we obtain poor feature extractor $\hbf_D$ after fine-tuning such that they are far from $\widehat{\hbf}$ due to inappropriate hyper-parameters: too large learning rate or too many iterations for stochastic gradient-based optimization.
As a result, the performance of the downstream task degrades because the model forgets the pre-trained weights to extract useful representations, and the downstream model overfits the downstream dataset.
To avoid this, pre-trained feature extractor $\widehat{\hbf}$ works as the explicit regularizer of $\hbf_D$.

Suppose the same notations and classification introduced in~\cref{sec:pre-training}.
Given pre-trained feature extractor $\widehat{\hbf}$,
we define the loss function for the downstream task with regularizer of $\hbf_D(.)$ as follows:
\begin{align}
  \min_{\hbf_D, \gbf_D} \widehat{L}_D(\hbf_D, \gbf_D) + \frac{\lambda}{N_D} \sum_{i=1}^{N_D} \Omega \left(\widehat{\hbf}(\xbf_i), \hbf_D(\xbf_i) \right),
  \label{eq:single-regularization}
\end{align}
where coefficient of regularization term $\lambda \in \Rbb_{\geq 0}$ and regularization function $\Omega: \Rbb^d \times \Rbb^d  \rightarrow \Rbb_{\geq 0}$.%
\footnote{Note that this formulation is called ``feature-based knowledge distillation'' in the distillation context~\citep{Gou2021IJCV}.}
For example, the $L_2$ distance between two representations is
\begin{align}
  \Omega \left( \widehat{\hbf}(\xbf), \hbf_D(\xbf) \right)
  = \left\| \widehat{\hbf}(\xbf) - \hbf_D(\xbf) \right\|_2.
\end{align}
Even if the dimensionalities of $\widehat{\hbf}(.)$ and $\hbf_D(.)$ are different, we still use this technique by adding an affine transformation to $\hbf_D$~\citep{Romero2015ICLR}.

As a similar formulation, we use the parameters of $\widehat{\hbf}$ for the regularizer of $\hbf_D$ rather than for only one representation \eqref{eq:single-regularization}.
Suppose pre-trained feature extractor $\widehat{\hbf}$ is modeled by a neural network with $J$ layers.
Let feature extractor's parameters be $\widehat{\boldsymbol{\theta}} = \{ \widehat{\wbf}^{(1)}, \widehat{b}^{(1)}, \ldots, \widehat{\wbf}^{(J)}, \widehat{b}^{(J)} \}$, where weights $\wbf$ and bias $b$.
Similarly, let $\boldsymbol{\theta}_{\hbf_D}$ be parameters in $\hbf_{D}$: $\boldsymbol{\theta}_{\hbf_D} = \left\{ \wbf^{(1)}_{\hbf_D}, b^{(1)}_{\hbf_D}, \ldots, \wbf^{(J)}_{\hbf_D}, b_{\hbf_D}^{(J)} \right\}$.
The counterpart of \cref{eq:single-regularization} with regularization term $\Omega(., .)$ of $\boldsymbol{\theta}_{\hbf_D}$ is defined as
\begin{align}
  \min_{\hbf_{D}, \gbf_{D}} \widehat{L}_{D}(\hbf_{D}, \gbf_{D}) +
  \lambda \Omega \left( \widehat{\boldsymbol{\theta}}, \boldsymbol{\theta}_{\hbf_D} \right).
  \label{eq:whole-regularization}
\end{align}
If we use $L_2$ distance as $\Omega(., .)$,
\begin{align}
  \Omega \left( \widehat{\boldsymbol{\theta}}, \boldsymbol{\theta}_{\hbf_D} \right)
  = \sum_{j=1}^J \left( \| \widehat{\wbf}^{(j)} - \wbf^{(j)}_{\hbf_D} \|_2 + \| \widehat{b}^{(j)} - b^{(j)}_{\hbf_D} \|_2 \right) .
  \label{eq:whole-l2-regularization}
\end{align}

As a special case of~\cref{eq:whole-l2-regularization}, we obtain $L_{2}$ regularization or ``weight decay'' in deep learning context when we set $\widehat{\boldsymbol{\theta}} = \mathbf{0}$ instead of pre-trained weights in $\widehat{\hbf}$:

\begin{align}
    \label{eq:l2-regularization}
    \Omega(\mathbf{0}, \boldsymbol{\theta}_{\hbf_D}) =
    \sum_{j=1}^J \left( \| \wbf^{(j)}_{\hbf_D} \|_2 + \| b^{(j)}_{\hbf_D} \|_2 \right) .
\end{align}
\citet{Li2018ICML} reported that \cref{eq:whole-l2-regularization} outperformed \cref{eq:l2-regularization} in a transfer learning scenario.
\cref{table:regularizations} categorizes studies that evaluated pre-trained feature extractors as explicit regularizers.

\subsubsection{Experimental procedure}
Since we use these regularizations to solve a downstream task, the evaluation procedure is the same as in~\cref{sec:pre-training-experimental-procedure}.

\subsubsection{Discussion}
These regularizations require more memory space, particularly \cref{eq:whole-l2-regularization} than \cref{sec:pre-training} because the number of parameters doubles for $\hbf_D$, which might make training infeasible, especially for large parameterized models, such as deep neural networks.
Therefore fine-tuning protocol with hyperparameter tuning is a more practical evaluation method than this explicit regularization.

A similar regularization term to~\cref{eq:whole-l2-regularization} can be obtained from PAC-Bayesian analysis~\citep{McNamara2017ICML}.
Through the lens of the PAC-Bayes analysis, $\widehat{\boldsymbol{\theta}}$ can be considered the prior of $\boldsymbol{\theta}_{\hbf_D}$.
If $\widehat{\boldsymbol{\theta}}$ is the good prior of $\boldsymbol{\theta}_{\hbf_D}$, \cref{eq:whole-l2-regularization} helps solve the downstream task.
In contrast, if we pick poor $\widehat{\boldsymbol{\theta}}$, the regularization hurts the optimization of the downstream task, making optimization unstable or leading to a poor feature extractor.

\subsection{Representation Learning for Dimensionality Reduction}
\label{sec:dimentionality}

\begin{table*}[t]
  \centering
    \begin{tabular}{lc}
    \toprule
      Evaluation task & References \\
    \midrule
    Reconstruction & \citep{Levy2014NeurIPS,Tosh2021JMLR} \\
    Statistics prediction & \citep{Schnabel2015EMNLP,Adi2017ICLR} \\
    Mutual information estimation & \citep{Hjelm2019ICLR,Tschannen2020ICLR} \\
    \bottomrule
  \end{tabular}
  \label{table:dimensionality-reduction}
  \caption{Summary of dimensionality reduction-based evaluation tasks for representation learning algorithms.}
\end{table*}

\begin{figure}[t]
  \centering
  \begin{subfigure}[b]{0.45\textwidth}
    \centering
      \includegraphics[width=\textwidth]{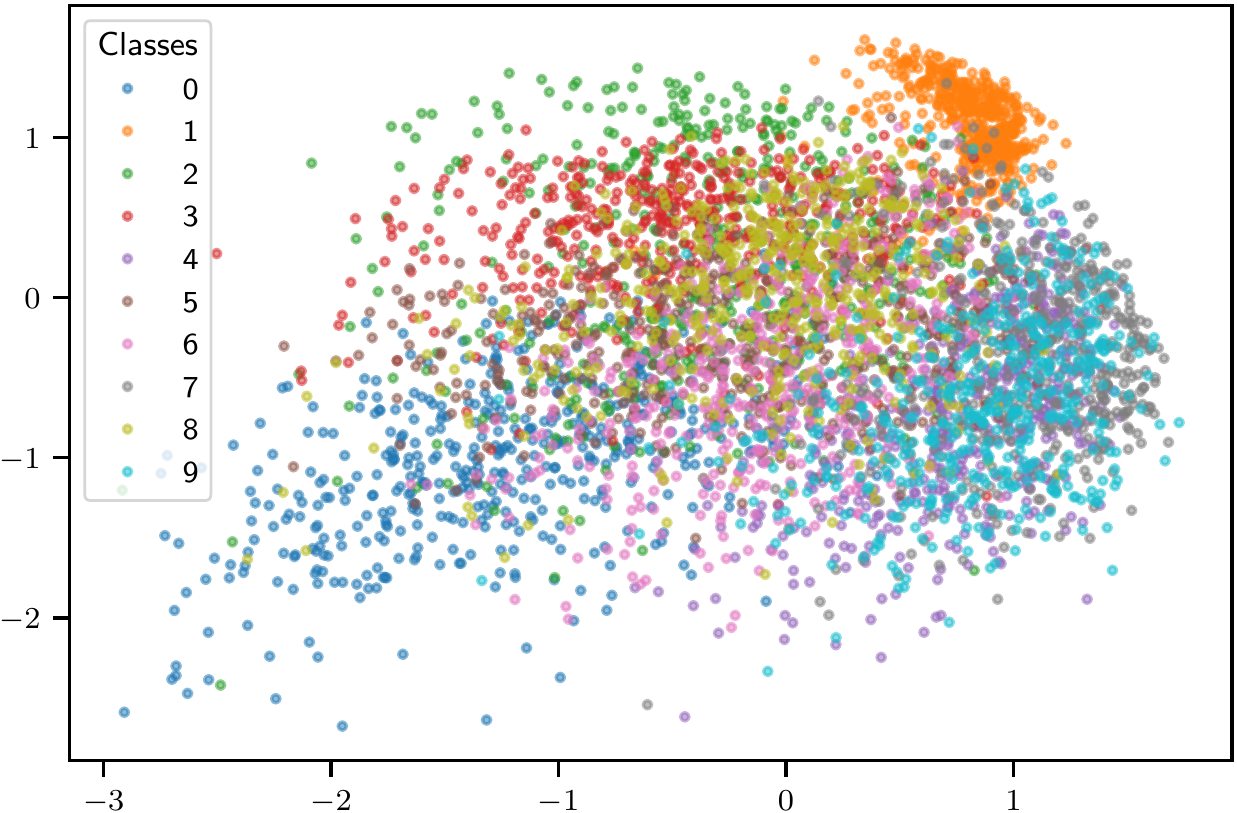}
      \caption{
        Visualization of intermediate representations extracted from MNIST by using an auto-encoder.
        The intermediate feature vectors are in $\Rbb^{2}$.
      }
      \label{fig:2d-representations-auto-encoder-mnist}
  \end{subfigure}
  \hfill
  \begin{subfigure}[b]{0.45\textwidth}
    \centering
      \includegraphics[width=\textwidth]{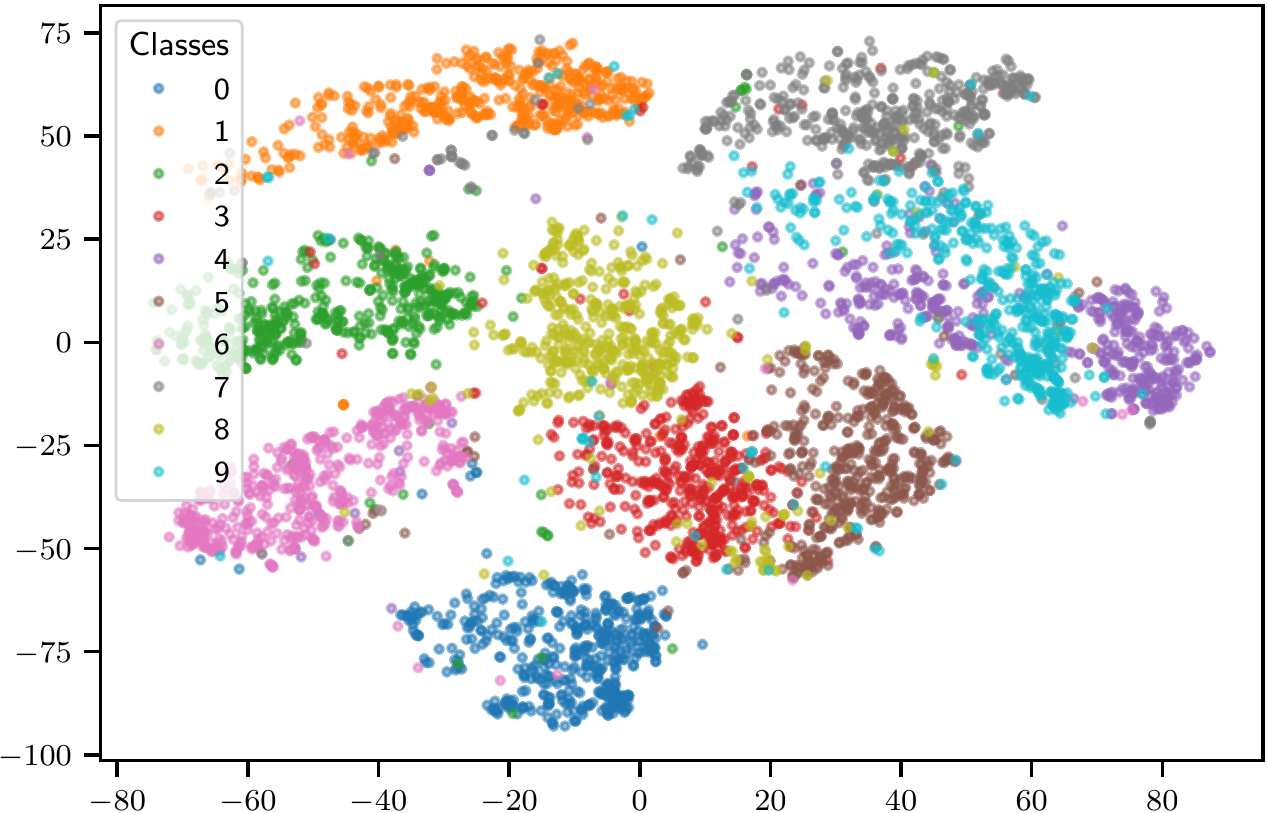}
      \caption{
        We apply $t$-SNE to the intermediate feature vectors of an auto-encoder, which are in $\Rbb^{16}$.
      }
      \label{fig:tsne-of-representations-auto-encoder-mnist}
  \end{subfigure}
  \caption{Two types of visualization of MNIST by using auto-encoders.
  \cref{sec:experimental-details} describes the details of visualization.
  }
\end{figure}

\begin{table*}
  \centering
  \begin{tabular}{lll}
    \toprule
    Reference &  Domain & Visualization algorithm \\
    \midrule
    \citet{Mikolov2013NeurIPS} & NLP & PCA \\
    \citet{Perozzi2014KDD} & Graph & Identity \\
    \citet{Tang2015WWW} & Graph & $t$-SNE \\
    \citet{Oord2018arXiv} & Speech & $t$-SNE \\
    \citet{Donahue2014ICML} & CV & $t$-SNE \\
    \citet{Kornblith2019CVPR}  & CV & $t$-SNE \\
    \citet{Islam2021ICCV}  & CV & $t$-SNE \\
    \citet{Erhan2010JMLR} & CV & Identity, ISOMAP \\
    \citet{Medina2020arXiv} & CV & $t$-SNE \\
    \citet{Chuang2020NeurIPS} & CV & $t$-SNE \\
    \citet{Wang2022ICLR}  & CV & $t$-SNE \\
    \citet{Tosh2021JMLR} & NLP & $t$-SNE \\
    \citet{Hashimoto2016TACL} & NLP & Identity \\
    \citet{Wang2020ICML}    & CV & Identity \\
    \citet{Kiros2015NeurIPS}& NLP & $t$-SNE \\
    \citet{Frosst2019ICML}  & CV, Synsetic & Identity, $t$-SNE \\
    \citet{Hadsell2006CVPR} & CV       & Identity \\
    \citet{Garg2020NeurIPS} & Synsetic & $t$-SNE \\
    \bottomrule
  \end{tabular}
  \label{table:visualization-only}
  \caption{Summary of used visualization algorithms to extracted feature representations.
  Identity means the dimensionaility of extracted by a feature extractor is $2$ or $3$; the authors visualized the extracted feature vectors directly without a visualization algorithm.}
\end{table*}

Dimensionality reduction~\citep{Espadoto2021IEEE} maps a raw data sample into a lower-dimensional space such that the mapped representation preserves important information from the original data sample.
For further details, please see general machine learning textbooks covering dimensionality reduction, for example, \citet[Chapter 23]{Ben-David2014},~\citet[Chapter 20]{Murphy2021Book}, and the survey paper~\citep{Espadoto2021IEEE} who review $44$ dimensionality algorithms and compare them empirically from the visualization perspective.%

The well known algorithms are independent component analysis~\citep{ICA2001book}, singular value decomposition, principal component analysis (PCA), LLE~\citep{Roweis2000Science}, Non-negative matrix factorization, $t$-distributed stochastic neighbor embedding ($t$-SNE)~\citep{Maaten2008JMLR}, and auto-encoder~\citep{Rumelhart1986Chapter}.
Representation learning works as dimensionality reduction when the dimensionality of extracted feature representation $d$ is smaller than the dimensionality of the original input $I$.
Indeed, matrix factorization-based dimensionality reduction algorithms are compared with unsupervised representation learning to extract feature vectors, especially discrete data such as word~\citep{Baroni2014ACL,Levy2015TACL} and node~\citep{Perozzi2014KDD,Tang2015WWW,Grover2016KDD}.
Similarly, a contrastive loss~\citep{Hadsell2006CVPR} proposed for dimensionaility reduction is used as a loss function of representation learning.

Data visualization can be viewed as a special case of dimensionality reduction when extracted feature representations are in $\Rbb^2$ or $\Rbb^3$, where a human can recognize features visually.
For example, for node representation learning, \citet{Perozzi2014KDD} showed feature vectors on $\Rbb^2$ that representation learning extracts directly from a graph dataset.
As a similar example, \cref{fig:2d-representations-auto-encoder-mnist} shows intermediate feature representation of an auto-encoder trained on MNIST dataset.
We also commonly apply a visualization algorithm to extracted feature representations $\big\{ \widehat{\hbf}(\xbf_i) \big\}_{i=1}^{N_D}$ rather than directly learn feature representations in $\Rbb^2$ or $\Rbb^3$.
For example, \citet{Mikolov2013NeurIPS} visualized word vectors with PCA, and \citet{Donahue2014ICML} visualized feature vectors extracted by convolutional neural networks with $t$-SNE~\citep{Maaten2008JMLR}.
As a similar example, \cref{fig:tsne-of-representations-auto-encoder-mnist} shows $t$-SNE visualization of feature representations extracted an auto-encoder.

\subsubsection{Experimental Procedure}
Suppose $R$ pre-trained feature extractors $\{ \widehat{\hbf}_r \}_{r=1}^R$ and downstream dataset $\Dcal_D$.
Extracted feature representations are denoted $\Dcal_{D, r} = \{ \widehat{\hbf}_r(\xbf_i) \}_{i=1}^{| \Dcal_D |}$.
We might not require the labels of $\Dcal_D$ depending on the evaluation metric.

\paragraph{Dimensionality reduction}
\begin{enumerate}
  \item Extract feature representations with each feature extractor $\widehat{\hbf}_r$ from $\Dcal_D$.
  \item Compare sets of extracted representations $\Dcal_{D, *}$ using evaluation metric for dimensionality reduction~\citep{Espadoto2021IEEE}.
\end{enumerate}

\paragraph{Visualization}
\begin{enumerate}
  \item Extract feature representations with each feature extractor $\widehat{\hbf}_r$ from $\Dcal_D$.
  \item If $d > 3$, apply a dimensionality reduction algorithm to the feature representations $\Dcal_{D, r}$ for visualization.
  \item Compare visualized features with scatter plot.
\end{enumerate}

\subsubsection{Discussion}
To our best knowledge, numerical evaluation of representation learning as dimensionality reduction is not performed commonly.
However, investigating what information is implicitly embedded in learned representations is actively performed to understand representation learning algorithms.
For example, given extracted sentence vectors, \citet{Adi2017ICLR} trained a simple model to predict a property of a sentence, e.g., sentence length, the existence of a word, or order of two words, in order to understand sentence feature extractors.
Intuitively, if the properties are embedded into extracted sentence vectors, the model can predict the properties accurately.
\cref{table:dimensionality-reduction} summarizes studies that evaluated representation learning algorithms from this perspective.

Numerical evaluation of visualization is challenging because it requires human evaluation, e.g., crowdsourcing~\citep{Borgo2018CGF}.
We believe that this is why representation learning papers showed only generated figures without numerical evaluation on visualization.
Another difficulty is that visualization algorithms generate different figures depending on their hyper-parameters.
For example, $t$-SNE algorithm gives largely different visualization results depending on its hyper-parameters~\citep{Wattenberg2016Distill}.
Hence we do not encourage evaluating representation learning algorithms using visualizations without careful the hyper-parameters tuning of visualization algorithms.
Another approach is to use label information if we access a label of each sample by following $t$-SNE's theoretical analysis~\citep{Arora2018COLT}.
In this case, we can use the similar evaluation protocol discussed in~\cref{sec:pre-training} with a simple classifier such as a linear classifier.
\cref{table:visualization-only} summarizes studies that evaluated representation learning algorithms with visualization.

\subsection{Representation Learning as Auxiliary Task}
\label{sec:auxiliary}

This evaluation perspective differs from the others.
We focus on a representation learning algorithm itself rather than pre-trained feature extractor $\widehat{\hbf}$.
Since representation learning attempts to learn generic feature representations from a dataset, the representation learning algorithm might improve another machine learning algorithm's performance by optimizing its loss and the loss of the downstream task simultaneously or cyclically, for example, in supervised~\citep{Islam2021ICCV,Frosst2019ICML}, semi-supervised learning~\citep{Zhai2019ICCV,Weston2008ICML}, few-shot learning~\citep{Gidaris2019ICCV}, domain adaptation~\citep{Carlucci2019CVPR}, and reinforcement learning~\citep{Oord2018arXiv}.
\cref{tab:auxiliary-summary} summarizes papers categorized into this evaluation perspective.

\begin{table*}
  {\footnotesize
  \centering
  \begin{tabular}{cccc}
    \toprule
    Reference & Main task & $L_{\mathrm{aux}}$ & Dataset of $L_{\mathrm{aux}}$ \\
    \midrule
    \citet{Oord2018arXiv} & Reinforcement learning & Contrastive~\citep{Oord2018arXiv} & S \\
    \citet{Mitrovic2021ICLR} & Reinforcement learning & Contrastive~\citep{Mitrovic2021ICLR} & S \\
    \citet{Chuang2020NeurIPS} & Reinforcement learning & Contrastive~\citep{Oord2018arXiv} & S \\
    \citet{Zhai2020arXiv} & Source task of transfer learning & Either of \citep{Gidaris2018ICLR,Dosovitskiy2014NeurIPS} & S, U \\
    \citet{Zhai2019ICCV} & Semi-supervised learning & Either of \citep{Gidaris2018ICLR,Dosovitskiy2014NeurIPS} & U \\
    \citet{Weston2008ICML} & Semi-supervised learning & Contrastive~\citep{Hadsell2006CVPR} &  U \\
    \citet{Gidaris2019ICCV} & Pre-training of few-shot learning & Either of \citep{Gidaris2018ICLR,Doersch2015ICCV} & S, U \\
    \citet{Carlucci2019CVPR} & Domain generalization, domain adaptation & Jigsaw~\citep{Noroozi2016ECCV} & S, Unl \\
    \citet{Luong2016ICLR} & Multi-task learning & Either of \citep{Kiros2015NeurIPS,Dai2015NeurIPS} & Unl \\
    \citet{Islam2021ICCV}  & Source task of transfer learning, Pre-training of Few-shot learning & Contrastive ~\citep{He2020CVPR} & S \\
    \citet{Frosst2019ICML} & Supervised learning & Soft nearest neighbor loss~\citep{Frosst2019ICML} & S \\
    \bottomrule
  \end{tabular}
  }
  \caption{
    The categorized experimental settings in \cref{sec:auxiliary}.
    For the dataset column,
    ``S'' stands for the same dataset as in the main task,
    ``U'' stands for an additional unlabeled dataset, and
    ``Uni'' stands for the union of the main task's dataset and additional unlabeled dataset.
  }
  \label{tab:auxiliary-summary}
\end{table*}

Suppose the same notations and classification formulation introduced in~\cref{sec:pre-training}.
Recall that we aim to learn a classifier that consists of feature extractor $\hbf_D$ and classification head $\gbf_D$ by minimizing supervised loss $L_D$, e.g., softmax cross-entropy loss~\eqref{eq:supervised-loss}.
Let $L_{\mathrm{aux}}$ be a representation learning's loss, e.g., mean squared loss \eqref{eq:auto-encoder-loss} for an auto-encoder-based representation learning algorithm and $\gbf_\mathrm{aux}$ be a representation learning specific projection head, e.g., the decoder of the auto-encoder model.
Suppose that the supervised and representation learning models share $\hbf_D$.
For example, $\hbf_D$ is a convolutional neural network, and loss-specific heads $\gbf_D$ and $\gbf_\mathrm{aux}$ are different fully-connected layers.
Formally, we optimize the following loss function to solve supervised loss $L_{\mathrm{D}}$ with representation learning loss $L_{\mathrm{aux}}$:
\begin{align}
  \min_{ \hbf_D, \gbf_D, \gbf_\mathrm{aux} } L_{\mathrm{D}}(\hbf_D, \gbf_D) + \beta L_{\mathrm{aux}}(\hbf_D, \gbf_\mathrm{aux}),
  \label{eq:RaaA-loss}
\end{align}
where pre-defined coefficient $\beta \in \Rbb_{\geq 0}$.
The feature extractor $\hbf_D$ can be released as a pre-trained feature extractor as the by-product of this supervised training~\citep{Zhai2020arXiv,Gidaris2019ICCV}.

\subsubsection{Experimental procedure}
Suppose $R$ representation learning algorithms.
For each representation learning algorithm, we optimize \cref{eq:RaaA-loss}.
We select the best one using the evaluation metric, such as validation accuracy for a classification task.

\paragraph{Variety of the dataset for representation learning}
There exist several ways to calculate representation learning loss $L_{\mathrm{aux}}$.
The simplest way is to use the same labeled dataset, $D_D$~\citep{Islam2021ICCV,Frosst2019ICML}.
Other ways are to use an additional unlabeled dataset~\citep{Luong2016ICLR} or the union of labeled and unlabeled datasets~\citep{Zhai2019ICCV,Weston2008ICML}.
When we use an unlabeled dataset, the unlabeled dataset can come from the same data distribution~\citep{Zhai2019ICCV} or a different data distribution~\citep{Luong2016ICLR}.

\subsubsection{Discussion}
This formulation can be seen as a multi-task learning~\citep{Caruana1997ML} whose tasks are the combination of a supervised task and a representation learning task.
Compared with the other evaluation perspectives, this evaluation is easy to tune hyper-parameters of the representation learning algorithm because we can search the hyper-parameters in the single training stage since pre-training in \cref{sec:pre-training} and regularization in~\cref{sec:regularization} require two-stage training: training the representation learning model and training the model of the downstream task.

\section{Theoretical Analyses of Representation Learning}
\label{sec:theoretical-analysis}

To understand the empirical success of representation learning, theoretical analysis of representation learning is as important as an empirical evaluation because it could yield novel representation algorithms and justify the empirical observation or techniques.
In this section, we mainly review theoretical analyses on unsupervised representation learning algorithms.

\subsection{Relationship between Representation Learning and Downstream Tasks}
\label{sec:theory-surrogate}

The majority of theoretical approach shows an inequality using the losses of representation learning and downstream tasks.
Loosely, given a feature extractor $\hbf$, the inequality is defined as
\begin{align}
  \label{sec:informal-bound}
  \widehat{L}_{D}(\hbf, \gbf_D) \leq \alpha \widehat{L}_{\mathrm{un}}(\hbf, \gbf) + \beta,
\end{align}
where $\alpha \in \Rbb_+$ and $\beta \in \Rbb$ are theorical analysis dependent terms.
This inequality implies that minimization of the loss of the downstream task $\widehat{L}_\mathrm{un}$ implicitly minimizes the loss of the representation learning task $\widehat{L}_D$.
The advantage of this type of inequality is that we can obtain generalization error bound in statistical learning theory~\citep{Ben-David2014} such as Rademacher complexity~\citep{Arora2019ICML} and PAC-Bayes-based complexity~\citep{Nozawa2020UAI}.

\subsubsection{Representation Learning Specific Analysis}

Most existing theoretical work specifies a representation learning task, especially the loss function.
We shall review the recent progress in this approach.

\paragraph{Contrastive loss with conditional independence.}
Since contrastive loss~\citep{Hadsell2006CVPR} is a state-of-the-art representation learning objective over the last decade~\citep{Mikolov2013NeurIPS,Oord2018arXiv}, theoretical analysis on the contrastive loss is actively performed.
Suppose we learn extractor $\hbf$ by minimzing a contrastive loss such as InfoNCE loss~\citep{Oord2018arXiv} on unsupervised dataset $D_{\mathrm{un}}$:
\begin{align}
  \widehat{L}_{\mathrm{cont}}(\hbf, \gbf) =
  - \frac{1}{M}
  \sum_{i=1}^M
  \ln
  \frac{
      \exp \left[ \mathrm{sim}(\gbf(\hbf(\xbf_i)), \gbf(\hbf(\xbf_i^+))) \right]
  }{
    \exp \left[ \mathrm{sim}(\gbf(\hbf(\xbf_i)), \gbf(\hbf(\xbf_i^+))) \right]
    + \displaystyle{\sum_{k \in [K]}}
      \exp \left[ \mathrm{sim} \left( \gbf(\hbf(\xbf_i)), \gbf \left( \hbf \left( \xbf_{(i, k)}^- \right) \right) \right) \right]
  },
  \label{eq:info-nce}
\end{align}
where $\mathrm{sim}(., .)$ is a similarity function, $\Rbb^O \times \Rbb^O \rightarrow \Rbb$, such as dot product or cosine similarity.
In \cref{eq:info-nce}, we need $K+1$ samples for each sample $\xbf_i$: positive sample $\xbf_i^+$ and $K$ negative samples $\{ \xbf_{(i, k)}^- \}_{k=1}^K$ that are randomly drawn from the dataset $D_\mathrm{un}$.

\citet{Arora2019ICML} showed the first theoretical analysis for contrastive unsupervised representation learning with hinge and logistic losses.
The key assumption of CURL is conditional independence on positive pair ($\xbf, \xbf^+$);
given a latent class $c$, the positive pair is sampled by $\xbf_i, \xbf_i^+ \sim p(\xbf \mid c)^2$.
Note that latent class $c$ cannot be observed directly, and the latent class is related to the supervised class in the (unseen) downstream task.
Under the same conditional independent assumptions, \citet{Nozawa2020UAI,Chuang2020NeurIPS} extended the results for non-vacuous generalization bound and the debiasing of negative sampling, respectively.
Similarly, \citet{Tosh2021ALT} weakened the assumption under $K=1$ and multi-view setting.
Instead of assuming a latent class, \citet{Tosh2021ALT} introduced a latent variable that makes two views conditionally independent.

\paragraph{Contrastive loss with data-augmentation.}
Since state-of-the-art unsupervised contrastive representation learning algorithms~\citep{Bachman2019NeurIPS,He2020CVPR,Caron2020NeurIPS,Chen2020ICML} generate positive pairs by applying data-augmentation to the same sample, the conditional independence assumption does not hold for these algorithms.
Relaxing the assumption is the main concern in theoretical analysis for contrastive unsupervised representation learning.
Note that this analysis is still useful in similar machine learning formulations such as metric learning and SU learning~\citep{Bao2018ICML} because this assumption holds naturally.

In \citet[Section 6.2]{Arora2019ICML}, a large number of negative samples $K$ in \cref{eq:info-nce} have a negative impact on the downstream tasks because we are more likely to draw a negative sample $\xbf^-$ whose latent class $c^-$ is the same as one of the positive pair.
On the contrary, recent contrastive representation learning algorithms~\citep{He2020CVPR,Chen2020ICML} reported that using large $K$ improves downstream performance in practice.
To deal with this regime, \citet{Nozawa2021NeurIPS} showed large $K$ is necessary to approximate the loss of the downstream task by extending the results by~\citet{Arora2019ICML} with the Coupon collector's problem and without the conditional independent assumption.
Similarly, \citet{Wang2022ICLR} proposed a theoretical analysis that deals with the large $K$ regime and incorporates data-augmentation explicitly.

As another line of theoretical approach, \citet{Mitrovic2021ICLR} showed conditions such that $\widehat{\hbf}$ generalizes to downstream tasks using a causal framework.
Intuitively, feature extractor $\widehat{\hbf}$ needs to i) distinguish finer-grained classes than ones of downstream tasks to generalize the unseen downstream tasks and ii) be invariant under the change of a style that is not a relevant part of an image for downstream tasks.
A contrastive loss~\citep{Dosovitskiy2014NeurIPS} and data-augmentation are justified to satisfy the two conditions, respectively.
\citet{HaoChen2021NeurIPS} proposed spectral graph-based analysis without the conditional independence assumption.
The authors consider a graph, where a node is an augmented data sample, and nodes are connected if they are generated from the data sample by data-augmentation.
Under this setting, the loss function of graph decomposition is similar to softmax-based contrastive loss~\citep{Oord2018arXiv}.
\citet{Jing2022ICLR} discussed the problem of the dimensional collapse of extracted feature representations by investigating the optimization dynamics of InfoNCE loss minimization with shallow linear neural networks.
\citet{Wen2021ICML} showed the importance of data-augmentation in contrastive learning on data generated by a sparse coding model with shallow ReLU neural networks.

\paragraph{Non-contrastive task}

Beyond contrastive representation learning, there exist analyses on different representation learning tasks.
For example, reconstruction-based representation learning tasks~\citep{Lee2021NeurIPS},
multi-task supervised representation learning for few-shot learning~\citep{Du2021ICLR},
self-training with data-augmentation~\citep{Wei2021ICLR}, vanilla language modeling~\citep{Saunshi2021ICLR}, and masked language modeling~\citep{Wei2021NeurIPS}.
\citet{Dubois2020NeurIPS} proposed Decodable Information Bottleneck to guarantee supervised representation learning performance.

\paragraph{Representation learning with auxiliary loss}
Theoretical analyses so far assume two stages of training as reviewed in \cref{sec:pre-training}.
For one-stage training as described in~\cref{sec:auxiliary}, theoretical analyses also are performed.
\citet{Le2018NeurIPS} showed the stability bound of a supervised loss with a linear auto-encoder used for the auxiliary task.
Similarly, \citet{Garg2020NeurIPS} proposed sample complexity bounds for unsupervised representation learning, which is a reconstruction-based task, and supervised learning losses.
In their analysis, intuitively, representation learning can work as a learnable regularizer to reduce the hypothesis size of the supervised model.
\citet{Maurer2016JMLR} gave sample complexity bounds for multi-task supervised learning and learning-to-learn via shared feature extractor $\hbf_D$ for all tasks.

\subsubsection{Representation Learning Agnostic Analysis}

A few studies analyze a predictor on downstream tasks given a pre-trained feature extractor $\widehat{\hbf}$.
The existing work showed the advantage of the frozen protocol with a linear classifier.
Concretely, \citet{Bansal2021ICLR} showed linear probing is more robust to the robustness of label corruption than fully-supervised way, and \citet{Kumar2022ICLR} showed linear probing generalizes better to out-of-domain dataset than fine-tuning after fitting on a downstream dataset.

\subsection{Optimality in Contrastive Representation Learning without Negative Samples}

Contrastive learning without negative samples (NCL) such as~\citep{Grill2020NeurIPS,Chen2021CVPR} is a novel type of unsupervised representation learning motivated by removing the negative samples in the contrastive loss \eqref{eq:info-nce}.
Even though NCL has collapsed minima: the outputs $\gbf(\fbf(.))$ are the same constant vector, empirically, it performs competitively well as a feature extractor for the downstream tasks.
\citet{Tian2021ICML} theoretically pointed out the importance of the stop-gradient technique and the existence of a projection head $\gbf$ to avoid such trivial solutions with linear neural networks.
\citet{Pokle2022AISTATS} showed that NCL has non-collapsed bad global minima, which means minima are far from ground truth weights under shallow feature extractor without projection head and synthetic data based on sparse coding with random mask data-augmentation setting.
On the other hand, contrastive unsupervised representation learning does not have such bad global minima.
If we appropriately set initialization and normalization for the weights in a feature extractor, the NCL converges to the ground truth~\citep[Theorem 3]{Pokle2022AISTATS}.
Unlike theoretical analyses for contrastive representation learning in \cref{sec:theory-surrogate}, these results do not give an inequality \eqref{sec:informal-bound};
revealing the relationship between the representation learning and downstream losses is worth interesting in exploring as a future direction.

\subsection{Relation to Other Metrics}
To understand what information is embedded by representation learning algorithms like \cref{sec:dimentionality}, researchers show the equivalence of optimization representation learning and other metrics.

For discrete data, especially natural language processing, information-theoretic analyses are actively proposed.
\citet{Levy2014NeurIPS} showed that skip-gram with negative sampling algorithm~\citep{Mikolov2013NeurIPS} implicitly decomposes a variant of point-wise mutual information (PMI) between word co-occurrence in the training corpus under assumptions, whose results are extended~\citep{Allen2019NeurIPS,Allen2019ICML}.
\citet{Qiu2018WSDM} extended the word embeddings results~\citep{Levy2014NeurIPS} to node representation learning algorithms.
Similarly, \citet{Hashimoto2016TACL} showed that word embedding algorithm could be formulated as matrix recovery, where the matrix is a PMI matrix of word co-occurrence counts,
and \citet{Arora2016TACL} showed similar results from a generative model perspective.
\citet{Kong2020ICLR} unified widely used algorithms in natural language processing from a mutual information perspective, including more recent models such as BERT~\citep{Devlin2019NAACL}.
Under the assumption of a generative model for documentations data, \citet{Tosh2021JMLR} showed contrastive representation learning could recover the topic posterior of a document.

Beyond discrete data, a relation between representation learning loss function and mutual information is discussed.
For example, contrastive loss function~\eqref{eq:info-nce} can be interpreted as a lower bound of mutual information~\citep{Oord2018arXiv,Hjelm2019ICLR,Poole2019ICML} or as an approximation of Hilbert-Schmidt independence criterion (HSIC)~\citep{Li2021NeurIPS}.
Note that tighter estimation of mutual information does not guarantee the generalization for the downstream task empirically~\citep{Tschannen2020ICLR,Tian2020NeurIPS}.
Complementary, \citet{Wang2020ICML} showed an interpretation of the role of a contrastive loss on hyper-sphere representation space.
\citet{Vincent2011NC} showned a connection between denoising autoencoder~\citep{Vincent2008ICML} and score matching~\citep{Hyvarinen2005JMLR} that is an efficient estimation method for probabilistic models.

\section{Conclusion and Future Directions}
\label{sec:conclusion}

Representation learning trains a feature extractor that automatically extracts generic feature representations from a dataset.
Unlike existing representation learning survey papers, we reviewed four evaluation methods of representation learning algorithms to understand the current representation learning applications.
We also reviewed theoretical work on representation learning.
We conclude this review by discussing the future directions based on Vapnik's principle~\citep{Vapnik2002book}.

A famous principle to solve a problem says
\begin{displayquote}[\citet{Vapnik2002book}][]
  When solving a given problem, try to avoid solving a more general problem as an intermediate step.
\end{displayquote}
Regarding the common evaluations in representation learning, representation learning seems to oppose Vapnik's principle.
For example, suppose a binary image classification: dog versus cat, as a downstream task.
We should not need a feature extractor that can distinguish the difference between Birman and Ragdoll, which are quite similar cat species, to solve the downstream task by following the principle.
However, we impose such ability on representation learning because it learns the generic feature extractor from a massive dataset for unseen downstream tasks.
In this sense, Vapnik's principle is inapplicable to representation learning.
Therefore we believe that we need a different metric to evaluate representation learning algorithms rather than the performance of a single downstream task, such as validation accuracy on ImageNet-1K.
One possible solution is to measure the averaged performance among diverse downstream tasks, such as VTAB~\citep{Zhai2020arXiv} for vision or SuperGLUE~\citep{Wang2019NeurIPSa} for language.
This idea can be generalized to modal-agnostic evaluation as discussed in \citet{Tamkin2021NeurIPSData}.

More pessimistically, solving the downstream task via representation learning, especially two-stage training, is less effective than solving the problem directly with comprehensive hyper-parameter tuning.
We expect the learned representations to capture redundant features to solve the downstream task, i.e., distinguishing between Birman and Ragdoll.
Such unnecessary expressiveness could hurt downstream tasks' performance.
In transfer learning terminology, a negative transfer might cause this ineffectiveness.
However, unsupervised representation learning has advantages compared with supervised learning, such as robustness to class imbalance~\citep{Liu2022ICLR} or generalization to unseen classes~\citep{Sariyildiz2021ICCV}.
Hence we reach the same future direction as in the previous paragraph: \textit{can we develop suitable evaluation metrics rather than only a single metric of a downstream task for representation learning?}

\section*{Acknowledgement}
This work is partially supported by Next Generation AI Research Center, The University of Tokyo.
We thank Han Bao and Yoshihiro Nagano for constructive discussion and for suggesting relevant work.

\appendix

\section{Experimental Settings}
\label{sec:experimental-details}
We trained a neural network with a fully connected hidden layer on the training dataset of MNIST.
We minimized a mean squared reconstruction error by using \texttt{AdamW}~\citep{Loshchilov2019ICLR} whose learning rate was initialized by $0.01$.
We adjusted the learning rate by using PyTorch~\citep{Paszke2019NeurIPS}'s \texttt{CosineAnnealingWarmRestarts}~\citep{Loshchilov2017ICLR} by epoch.
The mini-batch size was $500$, and the number of epochs was $20$.
As pre-processing, we normalized the input image.

We visualized feature vectors extracted by the hidden units on the validation dataset with {Matplotlib}~\citep{Hunter2007matplotlib}.
We used only $500$ samples per class to avoid showing too dense visualizations.
For \cref{fig:tsne-of-representations-auto-encoder-mnist}, we applied $t$-SNE implemented by {scikit-learn}~\citep{JMLR:v12:pedregosa11a} with its default hyper-parameters to the extracted feature representations whose dimensionality was $16$.
We developed the experimental codes on {Jupyter Notebook}~\citep{Kluyver2016ELPUB}.

{\scriptsize
\vskip 0.2in
\bibliography{reference}
\bibliographystyle{unsrtnat}
}

\end{document}